\documentclass[journal]{IEEEtran}
\usepackage[english]{babel}
\usepackage{graphicx}
\usepackage[small,bf]{caption}
\usepackage[centertags]{amsmath}
\usepackage{amsfonts}
\usepackage{amssymb}
\usepackage{amsthm}
\usepackage{ucs} 			% para codificacion unicode (utf8)
\usepackage[utf8x]{inputenc} 		% para codificacion unicode (utf8)
\usepackage{enumerate}
\usepackage{subfig}
\usepackage{ctable}
\usepackage{booktabs}
\usepackage{multirow}
\usepackage{nicefrac}

\usepackage{textcomp}

\usepackage{dsfont}

\newcommand\mydef{\mathrel{\overset{\makebox[0pt]{\mbox{\normalfont\tiny\sffamily def}}}{=}}}

\newcommand{\etal}{\textit{et al.}}
\newcommand{\overbar}[1]{\mkern 1.5mu\overline{\mkern-1.5mu#1\mkern-1.5mu}\mkern 1.5mu}

\definecolor{redStrong}{rgb}{.7,0,0}
\newcommand{\kn}{S_n}
\newcommand{\kw}{S_w}

\DeclareMathOperator*{\argmin}{arg\,min}
\DeclareMathOperator*{\argmax}{arg\,max}
\DeclareMathOperator*{\acoth}{acoth}

\newcommand{\K}{K}
\renewcommand{\ss}{\sigma^2}
\newcommand{\lm}{\lambda^2}
\renewcommand{\t}{\tau}
\newcommand{\I}{\mathbf{I}}
\newcommand{\E}{\mathbb{E}}
\newcommand{\oV}{\mathds{1} }
\renewcommand{\ln}{\log}
\newcommand{\mleA}{$\textsc{mle}_\text{avg}$}

\newcommand{\tr}{\text{tr}}
\newcommand{\mse}{\textsc{mse}}
\newcommand{\msec}{\text{\textsc{emse}}_1}
\newcommand{\crbd}{\textsc{crbd}}
\newcommand{\crbs}{\textsc{crbs}}
\newcommand{\crbsw}{\text{\textsc{crbs}}_w}
\newcommand{\crbsn}{\text{\textsc{crbs}}_{\text{n}}}

\newcommand{\ezzbw}{\textsc{ezzb}_w}

\newcommand{\snr}{\textsc{snr}}
\newcommand{\snrg}{\textsc{snr}}

\newcommand*\diff{\mathop{}\!\mathrm{d}}
\newcommand*\cond{\,\lvert\,}
 
\newcommand{\ba}{\mathbf{a}}

\newcommand{\bn}{\mathbf{n}}
\newcommand{\bx}{\mathbf{x}}
\newcommand{\bz}{\mathbf{z}}
\newcommand{\z}{z}
\newcommand{\bu}{\mathbf{u}}
\renewcommand{\u}{u}
\newcommand{\Np}{{N_p}}
\newcommand{\bomega}{\pmb{\omega}}

\newcommand{\Rdos}{\mathbb{R}^2}

\newcommand{\bS}{\mathbf{S}}
\newcommand{\bR}{\mathbf{R}}

\newcommand{\bP}{\mathbf{P}}
\newcommand{\bJ}{\mathbf{J}}
\newcommand{\bI}{\mathbf{I}}

\newcommand{\bdelta}{{\pmb{\delta}}}
\newcommand{\btau}{{\pmb{\tau}}}
\newcommand{\bSigma}{{\pmb{\Sigma}}}
\newcommand{\beps}{{\pmb{\epsilon}}}
\newcommand{\bphi}{{\pmb{\varphi}}}
\newcommand{\btheta}{{\pmb{\theta}}}

\newcommand{\nuC}{\mathcal{V}}

\newcommand{\Rk}{{\mathbb{R}^K}}

\ifCLASSINFOpdf
   \graphicspath{{fig/}}
\else
  \graphicspath{{fig/}}
\fi

\begin{document}

%\title{Fundamental performance bounds on multi-image registration}
\title{Fundamental Limits in Multi-image Alignment}


%\author{Cecilia~Aguerrebere, Andr\'es~Almansa, Julie~Delon, Yann~Gousseau and Pablo~Mus\'e\thanks{C. Aguerrebere is with the Department of Electrical and Computer Engineering, Duke University, Durham NC 27708, US (e-mail: cecilia.aguerrebere@duke.edu)}\thanks{A. Almansa and Y. Gousseau are with the Department of Signal and Image Processing, LTCI CNRS, T\'el\'ecom ParisTech, 75634 PARIS Cedex 13, France (e-mail: gousseau,almansa@telecom-paristech.fr).}\thanks{J. Delon is with MAP5 (CNRS UMR 8145), Universit\'e Paris Descartes, 75270 Paris Cedex 06 (e-mail: julie.delon@parisdescartes.fr)}\thanks{P. Mus\'e is with the Department of Electrical Engineering, Universidad de la Rep\'ublica, 11300 Montevideo, Uruguay (e-mail: pmuse@fing.edu.uy)}}

\author{Cecilia~Aguerrebere, Mauricio~Delbracio, Alberto Bartesaghi %~\IEEEmembership{Member,~IEEE,}
        and~Guillermo~Sapiro% <-this % stops a space
\IEEEcompsocitemizethanks{
%
%\IEEEcompsocthanksitem This work was partially funded by: 
%ONR, ARO, NSF, NGA, and AFOSR.
%
\IEEEcompsocthanksitem C. Aguerrebere, M. Delbracio and G. Sapiro are with the
Department of Electrical and Computer Engineering at Duke University. A. Bartesaghi is with the Laboratory of Cell Biology, Center for Cancer Research, National Cancer Institute, National Institutes of Health.

\noindent e-mail: \{cecilia.aguerrebere,~mauricio.delbracio,~guillermo.sapiro\}@duke.edu, bartesaghia@mail.nih.gov

}% <-this % stops a space
}%\thanks{Manuscript received April 19, 2005; revised September 17, 2014.}}

\maketitle


\begin{abstract}
The performance of multi-image alignment, bringing different images into one coordinate system, is critical in many applications with varied signal-to-noise ratio (SNR) conditions. A great amount of effort is being invested into developing methods to solve this problem. Several important questions thus arise, including: Which are the fundamental limits in multi-image alignment performance? Does having access to more images improve the alignment? Theoretical bounds provide a fundamental benchmark to compare methods and can help establish whether improvements can be made. In this work, we tackle the problem of finding the performance limits in image registration when multiple shifted and noisy observations are available. We derive and analyze the Cram\'er-Rao and Ziv-Zakai lower bounds under different statistical models for the underlying image. The accuracy of the derived bounds is experimentally assessed through a comparison to the maximum likelihood estimator. We show the existence of different behavior zones depending on the difficulty level of the problem, given by the SNR conditions of the input images. We find that increasing the number of images is only useful below a certain SNR threshold, above which the pairwise MLE estimation proves to be optimal. The analysis we present here brings further insight into the fundamental limitations of the multi-image alignment problem.
\end{abstract}


% \fi
\IEEEpeerreviewmaketitle

\section{Introduction}
Multi-image alignment consists in registering a group of images to a common reference.
\footnote{We use hereafter the terms image alignment and image registrationinterchangeably.} 
The multi-image alignment problem is ubiquitous in many fundamental image processing applications such as high dynamic range
imaging~\cite{debevec97,aguerrebere13ICCP}, super resolution~\cite{robinson2006statistical,robinson2009optimal,liu2014bayesian}, 
burst denoising~\cite{buades2009note} and burst deblurring~\cite{zhang2013multi,delbracio2015cvpr}. 
Indeed, this problem is of great importance for very different domains, such as biomedical imaging, astronomy and remote sensing, 
where due to physical or biological constraints the photographing system captures a series of unregistered and often noisy images. %We use hereafter the terms image alignment and image registration interchangeably.

%2- Many existent methods, some questions regarding perofrmance. Goal of this paper.
Various methods have been proposed for multi-image alignment~\cite{hardie97,sawhney98,govindu04,farsiu05,woods06,robinson2009optimal} and a great amount of effort is being invested to further improve their performance, mostly in applications that deal
with very low signal-to-noise ratio (SNR) conditions~\cite{li13,bartesaghi14,grant15,rubinstein15}.
Hence, an important question arises: Which are the fundamental limits on multi-image alignment performance? Theoretical performance bounds provide a fundamental benchmark to compare different methods 
and can help establish whether improvements can be made.
In this work, we tackle the problem of finding the performance limits in image registration 
when multiple shifted and noisy observations are available. 

%4. Importance of having performance bounds
Theoretical statistical performance bounds are of great interest and have been used in a wide variety of signal processing problems. 
One of the most widely used approaches, probably because of its simplicity, is the Cram\'er-Rao bound (CRB)~\cite{kay1993fundamentals}, 
which establishes a lower bound on the variance of any unbiased estimator of the parameter of interest. For instance, 
CRBs have been previously used to establish performance limits in pairwise image 
alignment~\cite{robinson04,pham05}, super-resolution~\cite{robinson2006statistical}, high dynamic range 
imaging~\cite{aguerrebere14} and image denoising~\cite{chatterjee10}, among others. Another example is the bound proposed by Ziv and Zakai~\cite{ziv1969some} and its extensions~\cite{seidman1970,chazan1975,bell1997extended}. 
They proposed to relate the mean squared error of the estimator to the probability of error in a binary detection problem, leading in general to tighter bounds than the CRB. Examples of the application of the Ziv-Zakai bound (ZZB) to practical problems can be found in pairwise image alignment~\cite{xu09} and time delay estimation~\cite{weiss1983fundamental,weinstein1984fundamental}, among others. Both the CRB and the ZZB will be computed here for the problem of multi-image alignment.

%3- Context in which we are going to tackle the problem: image translation
Image registration can easily become very complex with the kind of scene motions we face in real world scenarios. 
In this work, we focus on global translation which, despite being the most 
basic motion model, is of great interest because it is present in almost all applications. 
%
%6- Image Model: the u image
The considered motion model is thus given by
\begin{equation}
\z(\bx) = \u(\bx - \btau) + n(\bx),
\label{eq:introModel}
\end{equation}
where $\z(\bx)$ is the observed image at pixel position $\bx$, $\u$ is the underlying image, $\btau$ is the 2D translation 
vector and $n(\bx)$ is additive white Gaussian noise independent of $\u$. 

A fundamental aspect that has to be considered when computing a performance bound for the image alignment problem 
under Model~\eqref{eq:introModel}, is how to characterize the underlying image $\u$. 
Even if the parameter of interest is the shift vector $\btau$, assumptions have to be made about $\u$ and each assumption will lead to different performance bounds. 
For instance, $\u$ could be considered as deterministic, known or unknown, or as a realization of a known random process. 
%Different assumptions will lead to different performance bounds.

%7- Previous work on performance evaluation of pairwise image registration
Various performance bounds have been derived for the pairwise image alignment problem (i.e., registration between two images) 
assuming a deterministic \textit{known} underlying image. Examples of this are the CRB for translation estimation derived
by Robinson and Milanfar~\cite{robinson04}, the CRB for general parametric
registration introduced by Pham \etal{}~\cite{pham05} , and the ZZB derived by Xu \etal{}~\cite{xu09} for rigid pairwise registration including translation and rotation. 

%8- Previous work regarding evaluation of multi-image registration
Regarding multi-image alignment, a specific case was analyzed by Rais \etal{}~\cite{rais2014tight}, who computed the CRB
for the registration of a group of Earth satellite images that were uniformly
translated, i.e., all shifts are multiples of a single unknown value that needs to be estimated.
In~\cite{robinson2006statistical}, Robinson and Milanfar presented a thorough statistical performance analysis on super-resolution, of which multi-image registration is typically a major component. They studied translation estimation and image reconstruction jointly, thus assuming an \textit{unknown} underlying image. This work shed light on image super-resolution, giving important insight into which are the main bottlenecks for improving performance.
They derived bounds for the combined problem under two different assumptions for $\u$: the CRB assuming an unknown deterministic image and a Bayesian CRB assuming
a Gaussian prior for $\u$. 
In both cases, and assuming the considered images are aliasing free, the computed CRB for the multiple shifts estimation was independent of the number of available images. 

%8- TOA problem
It is interesting to remark that the problem of image translation estimation is closely related to the problem of time delay estimation 
of a signal observed at two or more spatially separated receivers~\cite{weiss1983fundamental,weinstein1984fundamental}. 
Indeed, our analysis follows and extends the results from~\cite{weinstein1984fundamental} to the case where multiple noisy versions of the same flat spectrum signal are observed, each with a different shift.

%9-A This work presents: Bounds
In this work, we derive and analyze various performance bounds for the multi-image alignment problem under two different models for the underlying image $\u$. First, we consider $\u$ to be deterministic and unknown. Under this image model, we compute the CRB and a Bayesian CRB assuming a generalized Gaussian prior for the shifts. Second, assuming a stochastic Gaussian model for the underlying image $\u$, we derive the CRB and the extended Ziv-Zakai bounds (EZZB).

A thorough analysis is conducted, which unveils the similarities between these seemingly different approaches. We find a \textit{per-region} behavior depending on the difficulty level of the problem, given by the SNR conditions. For certain SNR values, performance depends on the number of images. Also, it degrades dramatically below a given threshold, until reaching a region where the SNR is too low to enable alignment.

%9-A This work presents MLE
In order to assess the tightness of the computed bounds, we compare them to the alignment accuracy obtained by the maximum likelihood estimator (MLE). The MLE, besides being a widely used estimator, is known to be asymptotically efficient and also efficient for any number of observations in various problems~\cite{aguerrebere14}. A \textit{per-region} behavior depending on the SNR level, similar to the one predicted by the EZZB, is observed for the MLE as well. We find that all the computed bounds are very tight in very high SNR conditions, where the MLE achieves them and is thus efficient. For such high SNR, we find that the alignment performance only depends on the ratio between the energy of the image gradient and the noise level, and does not depend on the number of available images. Hence, for very high SNR, multi-image alignment can be performed in a pairwise fashion without losing information. 

However, this is not the case for low SNR where the performance shows a dependence on the number of images, until reaching a steady state error for extremely low SNR where alignment is not possible. 
The SNR values delimiting these regions, which are of particular importance in practice, are also derived and found to depend on the number of available images. Therefore, increasing the number of images is useful since, not only it improves the achievable performance, but it also shifts the SNR thresholds making alignment possible for a larger noise level range.

%10 - Organization of the article 
This article is organized as follows. Section~\ref{problemState} presents the statistical framework used to state multi-image alignment as a parameter estimation problem. Sections~\ref{sec:detImage} and~\ref{sec:randImage} are devoted to the computation and analysis of the different performance bounds, under the deterministic and stochastic image models respectively. Section~\ref{sec:boundsAnalysis} presents an analysis and comparison of all the computed bounds. The bounds accuracy is assessed in Section~\ref{sec:mle}. Section~\ref{sec:discussion} summarizes the conclusions. % presents a summary of the obtained results and perspectives. 

\section{Multi-image registration: an estimation problem}
\label{problemState}
In what follows, we present the image model used throughout the article for the derivation of the different performance bounds.
 Also, we introduce the performance indicators used to evaluate the translation estimators. 
Table~\ref{tab:notation} summarizes  the notation used in the article.
\begin{table}
\begin{center}
%\fbox{
%{\footnotesize
\begin{tabular}{ll} 
\toprule
$\z_i,\u$ & Images defined in continuous domain $\bx = [x,y] \in \Rdos$ \\
$\bz_i,\bu$ & Digital images sampled on discrete grid $m_r \times m_c$ \\
$\bu_x,\bu_y$ & Derivatives of $\bu$ in direction $x$ and $y$ \\
$\K$ & Number of unknown translations\\
%$\bz=\! [\bz_0^T,\dots,\bz_K^T]^T$ & Concatenation of all $(\K+1)$ observed digital images\\
$\btau_i $ & 2D translation vector $\btau_i = [\tau_{i_x},\tau_{i_y}]^T$ of image $i$ \\
%$\tau_{i_x},\tau_{i_y}$ & Horizontal and vertical translations of image $i$ \\
$\btau$ & Concatenation of $\K$ 2D translations \\
$\tilde{\z}_i,\tilde{u}$ & Fourier transform of images $\z_i$, $u$ \\
$\tilde{\bz}$ & Concatenation of $(\K+1)$  Fourier transforms $\tilde{\bz}_i$ \\
$\bomega$ & 2D Fourier spatial frequency $\bomega = [\omega_x,\omega_y]^T$\\
%$\omega_x,\omega_y$ & Horizontal and vertical Fourier spatial frequencies \\
$S(\bomega)$ & Power spectral density of 2D random process $\bu$\\
$\bJ_D,\bJ_S$ & Fisher information matrices\\
\textsc{mse} & Mean square error\\ 
\textsc{emse} & Expected mean square error\\
\textsc{snr} & Signal-to-noise ratio as defined by Eq.~\eqref{eq:SNR}\\
\textsc{crbd} & Cram\'er-Rao bound under deterministic image model Eq.~\eqref{trBound}\\
\textsc{bcrb} & Bayesian Cram\'er-Rao bound  (with shift prior) Eq.~\eqref{trBoundBayes}\\
\textsc{crbs} & Cram\'er-Rao bound under stochastic image model Eq.~\eqref{eq:crlbRand}\\
$\textsc{ezzb}_w$& Extended Ziv-Zakai bound (flat spectrum) Eq.~\eqref{eq:ezzbWS}\\
\bottomrule
\end{tabular}
%}
%\vspace{.5em} 
\end{center}
\caption{Summary of notation used in this article.}
\label{tab:notation}
\end{table}

\subsection{Image model}
Let us consider the image acquisition model:
\begin{equation}
\z_i(\bx) = \u(\bx - \btau_i) + n_i(\bx), \quad i=0,\dots,\K,
\label{eq:model}
\end{equation}
where $\z_i(\bx)$ is the observed $i$-th image at pixel position $\bx = [x,y]^T \in \Rdos$, $\u(\bx)$ is the underlying continuous image generating the noisy shifted observations, $\btau_i = [\tau_{i_x}, \tau_{i_y}]^T \in \Rdos$ is the 2D translation vector of frame $i$ with respect to the underlying image $\u$ (frame zero, $\btau_0 = 0$), and $n_i(\bx)$ is additive Gaussian noise assumed to be independent of $u$.

In practice, we do not have access to the continuous images but to a finite discretization of them. We will assume that all the images are band-limited and sampled according to the Nyquist sampling theorem. Regarding the finite observation support, we will additionally assume that the energy of the signal outside the observed sampling grid is negligible. These two assumptions guarantee an almost perfect interpolation of the \emph{continuous} images from the \emph{digital} ones.  Thus, under this ideal framework, we are able to compute image derivatives or image shifts (or any other linear operator) directly from the discrete samples.  Although we will omit the details for simplicity, all the considered operators could be computed via Fourier interpolation (e.g., using the \textsc{dft}).  Let us assume that the digital images are indexed into vectors of size $N_p = m_r \times m_c$ pixels, where $m_r$ and $m_c$ are the number of rows and columns respectively.

Let $\btau = [\btau_1^T,\dots,\btau_{\K}^T]^T \in \mathbb{R}^{2K}$ be the concatenation of all 2D unknown translations, and $\bz =\! [\bz_0^T,\dots,\bz_K^T]^T \in \mathbb{R}^{(K+1)\Np}$ be the concatenation of the $(K+1)$ observed images. The goal in multi-image alignment is then to estimate $\btau$ from $\bz$.
 
\subsection{Performance evaluation}
Let us call $\btheta$ the vector of parameters to be estimated, e.g. $\btheta = \btau$. Given any estimate $\hat{\btheta}(\bz)$ of $\btheta$, its performance can be measured through the error correlation matrix,
\begin{equation}
\mathbf{R}_\epsilon = \E_{\bz|\btheta} [\beps \beps^T ],
\end{equation}
where $\beps = \hat{\btheta} - \btheta$ is the error with respect to the real parameter value and $\E_{\bz|\btheta}[\cdot]$ is the expected value over the observed data distribution given $\btheta$. The fundamental limits on the estimation of $\btheta$ can be stated through the family of performance bounds which consider the parameter as an unknown \textit{deterministic} quantity and provide a limit on $\mathbf{R}_\epsilon$. Examples of this family are the Cram\'er-Rao~\cite{kay1993fundamentals}, Bhattacharyya~\cite{bhattacharyya1946some}, Barankin~\cite{barankin1949locally},  and Abel~\cite{abel1993bound} bounds, among others.

In some cases, prior information is known about $\btheta$. This motivates the development of the \textit{Bayesian} bounds, which model the parameter as a \textit{random variable} with a known \textit{prior} distribution, and give a limit on the expected error correlation matrix under the joint distribution of the data and the parameter
\begin{equation}
\overbar{\mathbf{R}}_\epsilon = \E_{\bz,\btheta} [\beps \beps^T ].
\end{equation}
Examples of Bayesian bounds are the Bayesian Cram\'er-Rao~\cite{trees1968detection}, the Ziv-Zakai~\cite{ziv1969some}, and the Weiss-Weinstein~\cite{weiss1985fundamental} bounds.

A more practical performance indicator is the mean squared error of the estimated parameters, which corresponds to the trace of the error correlation matrix. We refer hereafter as mean squared error (MSE) to the trace of $\mathbf{R}_\epsilon$ and expected mean squared error (EMSE) to the trace of $\overbar{\mathbf{R}}_\epsilon$.

In the following sections, we compute and analyze variants of two performance bounds for the multi-image alignment problem, the Cram\'er-Rao~\cite{kay1993fundamentals} and the Extended Ziv-Zakai lower bounds~\cite{bell1997extended}. The performance analysis is conducted under two different hypothesis for the unknown underlying image: $\bu$ is a deterministic unknown image (Section~\ref{sec:detImage}), and $\bu$ is a realization of a zero mean Gaussian random process with known covariance matrix (Section~\ref{sec:randImage}). Although the Gaussian model is over-simplistic~\cite{roth2005fields}, it is nonetheless interesting, not only because of its practicality, but also because it has proven to be very powerful for locally modeling natural images in several applications~\cite{elad2001fast,fransens2007optical,levin2009understanding,efrat2013accurate}.

\section{Performance bounds: deterministic image model}
\label{sec:detImage}
In this section, we assume that $\bu$ is an unknown deterministic digital image. 
We also assume that the noise in the digital observations $\bn$ has a diagonal covariance matrix $\sigma^2\mathbf{I}$. 
Notice that, even if the goal of multi-image registration is to estimate $\btau$ and not $\bu$, 
the latter is unknown and needs to be accounted for in the analysis. 
This kind of parameters, whose estimation is not of direct interest but because they are related to the analysis have to be accounted for,
are commonly referred to as nuisance parameters~\cite{vanTrees2013detection}.
Hence, the  parameter vector becomes $\btheta=[\bu^T, \btau^T]^T$, where we are only interested in estimating $\btau$ from
the $(\K+1)$ noisy observed images $\bz$. 
 
\subsection{Cram\'er-Rao lower bound: deterministic image model}
The performance of any unbiased estimator $\hat{\btheta}(\bz)$ of $\btheta$ is bounded by the CRB~\cite{kay1993fundamentals}
\begin{equation}
\mathbf{R}_\epsilon \geq \mathbf{J}^{-1},
\label{CRB}
\end{equation}
where $\mathbf{J}$ is the Fisher information matrix (FIM) given by
\begin{equation}
\{\mathbf{J}\}_{i,j} = -\E_{\bz|\btheta} \left [ \frac{\partial^2 \ell(\bz ; \btheta )}{\partial \theta_i\partial \theta_j} \right ],
\label{eq:FIM}
\end{equation}
and $\ell(\bz ; \btheta ) = \log ( p(\bz;\btheta))$ is the logarithm of the likelihood function. The FIM in this case can be expressed as
\begin{equation}
\bJ_D = 
      \begin{bmatrix}
	\mathbf{J}_{\bu\bu} & \mathbf{J}^T_{\bu\btau} \\      
        \mathbf{J}_{\bu\btau}  & \mathbf{J}_{\btau \btau} 
      \end{bmatrix},
      \label{eq:fimOrig}
\end{equation}
where the term $\mathbf{J}_{\bu\bu}$ captures the information provided by the image only, the term $\mathbf{J}_{\btau \btau}$ captures the available information of the translations and $\mathbf{J}_{\bu\btau}$ represents the information provided by the intercorrelation between $\bu$ and $\btau$. Using the block matrix inversion principle~\cite{matrixCookbook2012}, the inverse of $\mathbf{J}$ can be expressed as
\begin{equation}
\bJ_D^{-1} = 
      \begin{bmatrix}
	\mathbf{S}_{\bu}^{-1} & \mathbf{J}^{-1}_{\bu\bu}\mathbf{J}_{\bu\btau}\mathbf{S}_{\btau}^{-1} \\      
        \mathbf{S}^{-1}_{\t}\mathbf{J}_{\bu\btau}^T\mathbf{J}_{\bu\bu}^{-1}  & \mathbf{S}_{\btau}^{-1} 
      \end{bmatrix},
\label{eq:invJd}      
\end{equation}
where $\mathbf{S}_{\btau}$  and $\mathbf{S}_{\bu}$  are the Schur complements of the submatrix  regarding $\btau$ and $\bu$ respectively, namely,
\begin{align}
\mathbf{S}_{\btau} &= \mathbf{J}_{\btau\btau}  - \mathbf{J}_{\bu\btau}^T \mathbf{J}_{\bu \bu}^{-1} \mathbf{J}_{\bu\btau}, \\
\mathbf{S}_{\bu} &= \mathbf{J}_{\bu\bu}  - \mathbf{J}_{\bu\btau} \mathbf{J}_{\btau \btau}^{-1}\mathbf{J}^T_{\bu\btau}. 
\end{align}
It can be shown that for multi-image registration, $\mathbf{S}_{\btau}^{-1}$ is given by (see Appendix~\ref{Ap:detCRB})
\begin{equation}
\mathbf{S}_{\btau}^{-1} = \sigma^2 [ \mathbf{I}_{\K} + \oV \oV^T] \otimes \mathbf{Q}^{-1},
\end{equation}
where $\I_{\K}$ is the identity matrix of size $\K \times \K$, $\oV$ is a vector of ones of size $\K$, $\otimes$ is the Kronecker product between matrices,
\begin{equation}
\mathbf{Q} = 
      \begin{bmatrix}
	\bu_x^T\bu_x & \bu_x^T\bu_y \\      
        \bu_x^T\bu_y  & \bu_y^T\bu_y  
      \end{bmatrix},
\end{equation}
and $\bu_x$, $\bu_y$ are the derivatives of the latent image $\bu$ in the horizontal and vertical directions respectively.

Equation~\eqref{CRB} gives a bound on the covariance matrix of any unbiased estimator of $\btheta$. Therefore, from~\eqref{CRB} and~\eqref{eq:invJd}, the MSE of the estimated translations is bounded by the trace of $\mathbf{S}_{\btau}^{-1}$~\cite{nehorai00},
\begin{align}
\mse &= \tfrac{1}{2\K} \sum_{j=1}^{\K} \E[(\tau_{j_x} \!- \hat{\tau}_{j_x})^2 \!+ (\tau_{j_y} \!- \hat{\tau}_{j_y})^2] \label{eq:mse}\\
&\ge \tfrac{1}{2\K} \tr(\mathbf{S}_{\tau}^{-1})\\
&= \sigma^2 \frac{(\bu_x^T\bu_x + \bu_y^T \bu_y)}{(\bu_x^T\bu_x) (\bu_y^T \bu_y) - (\bu_x^T \bu_y)^2}.
%= 2\sigma^2 \frac{(u_x^Tu_x + u_y^T u_y)}{(u_x^Tu_x) (u_y^T u_y) - (u_x^T u_y)^2}.
\end{align}
Hence, we define the CRB under a deterministic image model (CRBD) as,
\begin{equation}
\crbd \mydef  \sigma^2 \frac{(\bu_x^T\bu_x + \bu_y^T \bu_y)}{(\bu_x^T\bu_x) (\bu_y^T \bu_y) - (\bu_x^T \bu_y)^2}.
\label{trBound}
\end{equation}
According to the CRBD, the registration error is proportional to the noise level and inversely proportional to the energy of the gradient.
A similar result is presented by Robinson and Milanfar~\cite{robinson2006statistical}, who derived the CRB for the super-resolution problem. Multi-image registration can be seen as a particular case of the super-resolution problem, where the under-sampling operator is equal to the identity matrix. 

\smallskip

\noindent \textbf{Performance independence of $\K$}. An unexpected result is that the bound~\eqref{trBound} does not depend on the number of images $K$. This means that this fundamental limit of multi-image registration performance is the same for a set of 2 or any number $K$ of images. Nevertheless, unlike stated in~\cite[Ap. III]{robinson2006statistical}, this does not imply that registration can be done pairwise without loss of information. The CRB gives a lower bound on performance but it does not ensure the existence of an efficient estimator that reaches this bound. In practice, depending on the problem, the CRB may or may not be tight. Hence, the independence of the CRB of the number of images $\K$, does not imply that registration can be done pairwise without loss of information.

As it will be shown experimentally in Section~\ref{sec:mle}, for the multi-image registration problem, the CRBD is tight in high SNR conditions, where we observe indeed that registration can be done pairwise without loss of accuracy. However, it is not necessarily tight in low SNR conditions. In that case, there are other bounds, which are dependent on $\K$, that are closer to the actual performance estimators can achieve.

\smallskip

\noindent \textbf{Case with known underlying image.} The bound on~\eqref{trBound} corresponds to the translations estimate error when the real image $\bu$ is unknown, which is usually the case in practice. In previous works~\cite{robinson04,pham05}, however, the CRB has been computed for the pairwise image registration problem when the only unknown parameters are the shift values. 

In that case, the FIM for the multi-image registration problem \eqref{eq:fimOrig} simplifies to
\begin{equation}
\bJ_{D_\text{kn}} = \mathbf{J}_{\btau \btau} = \frac{1}{\sigma^2} \bI_K \otimes \mathbf{Q},
\end{equation}
and the CRB for the case where $\bu$ is known becomes,
\begin{align}
\crbd_\text{kn} &\mydef  \tfrac{1}{2K} \tr (\mathbf{J}^{-1}_{\btau \btau})\\
    &=  \frac{\sigma^2}{2} \frac{(\bu_x^T\bu_x + \bu_y^T \bu_y)}{((\bu_x^T\bu_x) (\bu_y^T \bu_y) - (\bu_x^T \bu_y)^2)}= \frac{\crbd}{2}.
%   &= \frac{\crbd}{2}.
\label{trBound_uknown}
\end{align}
Therefore, the MSE bound, assuming the underlying image $\bu$ is known, is half that of the case when $\bu$ is unknown. 
When $\bu$ is known, and the first image in the set is assumed to be aligned (i.e., $\btau_0 = 0$), all the other images
can be pairwise aligned to the known reference. Indeed, in that case, the different observed images are conditionally
independent given the known underlying image. Hence, there is no gain in using the rest 
of the images for estimating the translation of one image.
Therefore, the limiting factor in the pairwise alignment is the noise in one image.

On the other hand, when $\bu$ is unknown, the bound doubles. This may represent a best case scenario where the limiting factor is twice the noise, 
corresponding to the pairwise alignment of two noisy images.

\subsection{Bayesian Cram\'er-Rao with prior on shifts}
\label{ssec:BCRBshifts}
A natural question that arises after finding that the fundamental performance limit given by the CRBD~\eqref{trBound} does not depend on the number of images, is whether this limit can be improved if some prior information about the shifts is  known. Intuitively, having more images could improve the alignment performance in the case that $\bu$ is unknown. Let us imagine the case of an algorithm that uses an estimation of the latent image $\bu$ to estimate the shifts. One could expect that the estimation of $\bu$ could be improved by having more images, for example by reducing the noise, and thus leading to a better estimate of the shifts. 

Hence, the question is what happens if the motion estimation can be improved by including prior knowledge on the shift vectors.
A typical assumption is that the shifts are independent and drawn from a uniform distribution within a limited range. Nevertheless, in some particular applications (e.g., in microscopy or in burst photography), each shift vector depends on the previous ones so modeling the motion as a random walk might be more accurate. In this work, we restrict the analysis to the case where the shifts are independent.

A Bayesian version of the CRB bound can be computed to include prior information on the unknown parameters. The Bayesian Cram\'er-Rao bound (BCRB) gives a lower bound on the expected error correlation matrix under the joint data and parameter distribution
\begin{equation}
\overbar{\mathbf{R}}_\epsilon \geq \mathbf{J}_B^{-1},
\label{BCRB}
\end{equation}
where $\bJ_B$ is the Bayesian Fisher information matrix given by
\begin{equation}
\{\mathbf{J}_B\}_{i,j} = -\E_{\bz,\btheta} \left [ \frac{\partial^2 \ell(\bz , \btheta )}{\partial \theta_i\partial \theta_j} \right ],
\label{eq:BFIM}
\end{equation}
and $\ell(\bz , \btheta ) = \log ( p(\bz,\btheta))$ is the logarithm of the joint likelihood function.

%\medskip
\noindent \textbf{Generalized Gaussian prior on shifts.} Let us consider a centered generalized Gaussian prior distribution for each component of $\btau$. This family of densities, indexed by the parameters $c>0$ and $\delta>0$, is given by
\begin{equation}
p(\tau; c,\delta) = \frac{c\,\eta(\delta,c)}{\Gamma(1/c) \exp{( -\eta^c(c,\delta)|\t|^c})},
\label{eq:GenGaus}
\end{equation}
where
\begin{equation}
\eta(\delta,c) = \frac{1}{\delta} \left (\frac{\Gamma(3/c)}{\Gamma(1/c)} \right)^{1/2},
\end{equation}
and $\Gamma$ denotes the gamma function. The case of $c=2$ corresponds to the Gaussian density, while the distribution
approaches the uniform density with variance $\delta^2$ as $c \to \infty$. 

%\medskip

Given $(\K+1)$ independent samples following model~\eqref{eq:model}, and assuming that $\bu$ is an unknown deterministic image and $\btau$ is a random variable following the generalized Gaussian prior~\eqref{eq:GenGaus}, the Bayesian FIM is given by (see Appendix~\ref{app:BCRB})
\begin{equation}
\bJ_B = \bJ_D + \mathbf{J}_p,
\end{equation}
with 
\begin{equation}
\mathbf{J}_p = 
      \begin{bmatrix}
	0 & 0 \\      
        0  & \tfrac{1}{\lm} \mathbf{I}_{2\K}
      \end{bmatrix},
\end{equation}
where $\lm = \frac{\delta^2 \Gamma^2(1/c)}{c^2\Gamma(3/c)\Gamma(2-1/c)}$ and $\bJ_D$ is the FIM given in $\eqref{eq:fimOrig}$. 
Therefore, including a prior on the translations adds the term $\mathbf{J}_p$ to the classical FIM given in \eqref{eq:fimOrig}. 

Then, the Schur complement of the submatrix regarding $\btau$, becomes (see Appendix~\ref{app:BCRB})
\begin{align}
\bar{\mathbf{S}}_{\btau} &= \tfrac{1}{\ss}\left (  \mathbf{I} - (K+1) \oV \oV^T \right ) \otimes \mathbf{Q} + \tfrac{1}{\lm} \mathbf{I},
\end{align}
and
\begin{align}
\bar{\mathbf{S}}_{\btau}^{-1} &= \mathbf{I} \otimes \left ( \tfrac{1}{\ss} \mathbf{Q} + \tfrac{1}{\lm} \I \right )^{-1} \label{Sinv} \\
                        &\phantom{=}+ \oV \oV^T \otimes \lambda^2 \left ( (\K\! +\! 2) \I + \tfrac{\lm}{\ss} \mathbf{Q} + (\K\! +\!1) \tfrac{\ss}{\lm} \mathbf{Q}^{-1} \right )^{-1}.\nonumber
\end{align}
The EMSE of the translations under the given prior is then lower bounded by
%$$\textsc{emse} = \E_\btau  \left[ \frac{1}{2\K} \sum_{j=1}^{\K} \E[(\tau_{j_x} \!- \hat{\tau}_{j_x})^2 \!+ (\tau_{j_y} \!- \hat{\tau}_{j_y})^2]  \right ]$$ under the given prior is bounded by
\begin{equation}
\textsc{emse} \ge \textsc{bcrb} \mydef \tfrac{1}{2\K} \tr(\bar{\mathbf{S}}_{\btau}^{-1} ).
\label{trBoundBayes}
\end{equation}

%\smallskip

\begin{figure}
\centering
\includegraphics[width=0.7\linewidth]{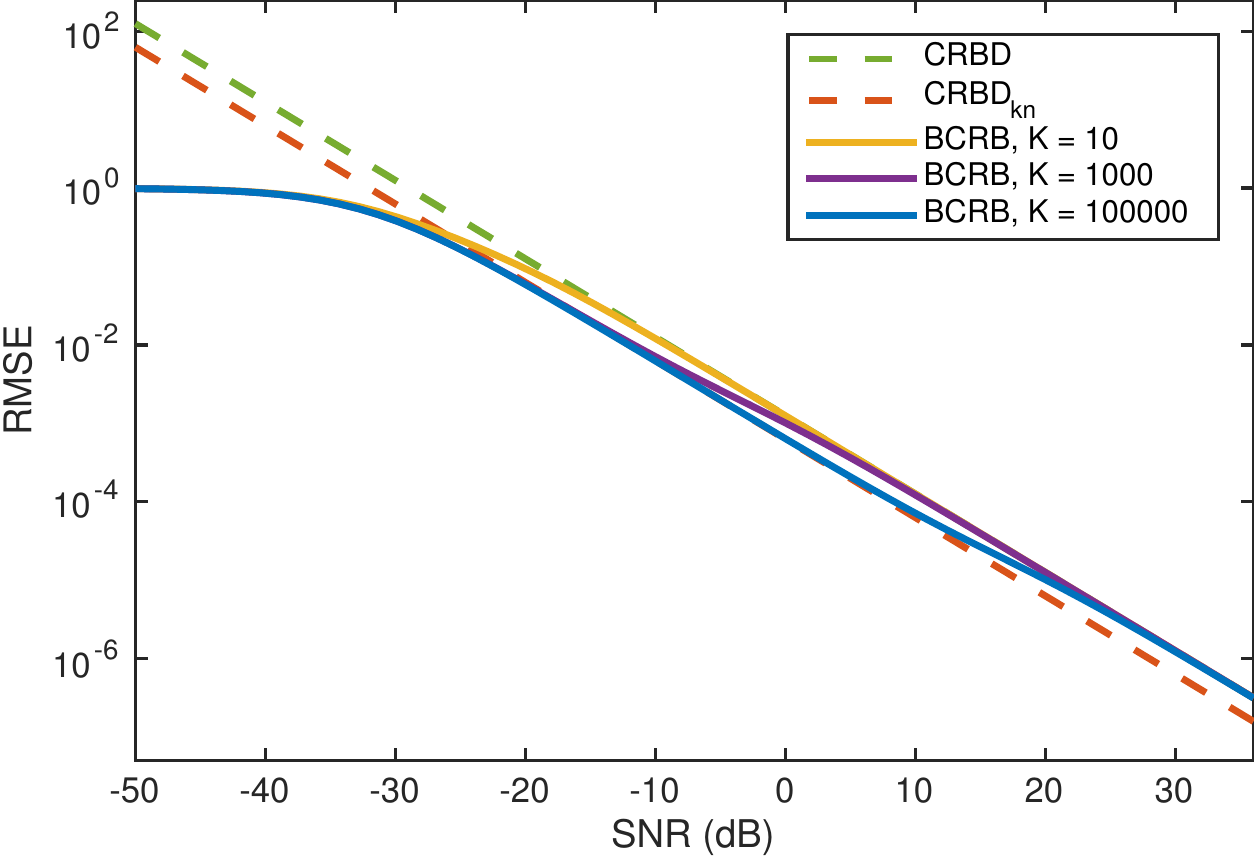}
\caption{Comparison of the CRBD (both for $\bu$ known and unknown)
and the BCRB for different number of images $\K$ with a Gaussian prior ($c=2$) with $\delta=1$ on the shifts. }%. \textbf{Left:} Wide SNR range. \textbf{Right:} Zoom of the central part. }
\label{fig:comparCRBs}
\end{figure}

%
As a first observation, let us point out that adding prior information on the translations makes the bound dependent on the 
number of images $\K$. Figure~\ref{fig:comparCRBs} shows a comparison of the CRB ($\bu$ known and unknown)
and the BCRB for different number of images $\K$ with a Gaussian prior ($c=2$) with $\delta=1$. 
Note that both bounds are very similar for high SNR values. 
This means that if the SNR is high enough, there is no gain in having prior knowledge about the shifts. 
However, for low enough SNR and large enough $\K$, the prior bounds the errors on the shift estimates and
the BCRB is below the CRBD and approaches the bound for a known image $\text{CRBD}_{kn}$ until it reaches a steady-state
value equal to the variance of the prior.

Notice that the example shown in Figure~\ref{fig:comparCRBs} corresponds to a pretty tight prior ($\delta=1$), 
meaning that an accurate interval for the shifts is known a priori. 
Remarkably, even under this seemingly favorable condition, the reduction of the BCRB is observed only for a very low 
SNR range and for a very large number of images. This suggests a very limited impact of this shift prior in practice,
being useful only for very low SNR conditions, a tight prior of the shifts interval, and a very large number of images. 

The generalized Gaussian prior approaches the uniform distribution when $c\to \infty$. 
Thus, for any fixed $\delta$, $\lambda \to 0$ as the shift prior approaches a uniform distribution. 
Hence, the prior information becomes irrelevant and the performance is bounded by the 
CRBD, which is independent of $K$. Of course, this does not mean that having more images does not help
for estimating the shifts. As previously mentioned, if the CRBD is overoptimistic and cannot be attained, a tighter bound may still exist, that does depend on the number of images.

\section{Performance bounds: stochastic image model}
\label{sec:randImage}
In this section, we consider a zero-mean Gaussian stochastic model for the underlying unknown image $\bu$. 
As stated before, our goal is to estimate the $K$ shifts between every pair of $\K+1$ images given by \eqref{eq:model}, or 
equivalently in the Fourier domain, from
\begin{equation}
\tilde{z}_i(\bomega) = \tilde{u}(\bomega) e^{-i \bomega \cdot \btau_i} + \tilde{n}_i(\bomega), \quad i=0,\ldots,\K,
\end{equation}
where \texttildelow\, denotes 2D image Fourier transforms,  $\bomega = [\omega_x,\omega_y]^T$ represents the 2D Fourier spatial frequency and $\cdot$ denotes the inner product operation.

We now assume that the  signal samples $\bu$ are drawn from a stationary zero-mean Gaussian process with spectral density $S(\bomega)$.  
The additive noise is modeled by the zero-mean Gaussian process $\bn_i$ with spectral density $N(\bomega)$, assumed to be independent of the underlying signal $\bu$. 

The observed digital images $\bz_i$ can be converted into the Fourier domain $\tilde{\bz}_i$ by applying the 2D \textsc{dft}. In practice, since the input images are real, the complex Fourier coefficients have Hermitian symmetry, where two of the four quadrants fully determine $\tilde{\bz}_i$. Here, we arbitrarily choose to work with the positive values of $\omega_y$ and the complete range for $\omega_x$ (i.e., first and second quadrants of the 2D  \textsc{dft}). Hence, we will only consider the complex Fourier coefficients corresponding to frequencies $\bomega_{l_x,l_y} =[\omega_{l_x},\omega_{l_y}]^T$ with $\omega_{l_x} = \frac{2\pi l_x}{m_c}, l_x = -\frac{m_c}{2},\dotsc,\frac{m_c}{2}$ and $\omega_{l_y} = \frac{2\pi l_y}{m_r}, l_y = 0,\dotsc,\frac{m_r}{2}$.
In addition, we will assume that the Fourier coefficients of $\bu$ are uncorrelated at the considered spatial frequencies. 

Let $l(l_x,l_y)=1,\dotsc,M$, with $M=m_c + \frac{m_r}{2} + 2$, index all the considered 2D frequencies $\bomega_l$. The Fourier transform of the ($\K+1$) observed images can be arranged into a vector
\begin{equation}
\begin{split}
\tilde{\bz} = [\tilde{z}_0(\bomega_1),\tilde{z}_1(\bomega_1),\ldots,\tilde{z}_K(\bomega_1),\ldots,\\\tilde{z}_0(\bomega_M),\tilde{z}_1(\bomega_M),\ldots \tilde{z}_K(\bomega_M)]^T.
\label{eq:zFourier}
\end{split}
\end{equation}

Under Gaussian assumptions for the noise and the underlying image, $\tilde{\bz}$ follows a complex Gaussian distribution with zero mean and covariance matrix
\begin{equation}
\bSigma = \E[\tilde{\bz} \tilde{\bz}^H] =
\begin{bmatrix}
\bSigma_\btau (\bomega_1) & 0 & \ldots & 0 \\
0 & \bSigma_\btau (\bomega_2)& \ldots & 0 \\
\vdots & \vdots & \ddots & \vdots \\
0 & 0 & \ldots & \bSigma_\btau(\bomega_M)
\end{bmatrix},
\label{eq:Kdef}
\end{equation}
where each matrix $\bSigma_\btau(\bomega)$ has size $(K+1)\times(K+1)$ and is composed by
\begin{equation}
{\small
\setlength{\arraycolsep}{2pt}
\bSigma_\btau(\bomega)\!=\!\!\!
\begin{bmatrix}
S(\bomega) \!+\! N(\bomega) & \!\!\!S(\bomega) e^{-i \btau_1 \cdot \bomega} & \!\!\!\ldots\!\!\! & S(\bomega) e^{-i \btau_K  \cdot \bomega} \\
S(\bomega) e^{i \btau_1 \cdot  \bomega} & \!\!\!S(\bomega) \!+\! N(\bomega) & \!\!\!\ldots\!\!\! & S(\bomega) e^{i (\btau_1 \!- \btau_K)  \cdot \bomega} \\
\vdots & \!\!\!\vdots & \!\!\!\ddots\!\!\! & \vdots \\
S(\bomega) e^{i  \btau_K \cdot  \bomega} & \!\!\!S(\bomega) e^{-i(\btau_1 \!- \btau_K)  \cdot \bomega} & \!\!\!\ldots\!\!\! & S(\bomega) \!+\! N(\bomega)
\end{bmatrix}.
}
\end{equation}

\subsection{Cram\'er-Rao lower bound: stochastic image model}
In order to compute the CRB for the shifts estimation in the multi-image alignment problem under model~\eqref{eq:model}, we first compute the corresponding FIM matrix. For the considered complex Gaussian process $\hat{\bz}$, it is given by \cite[Ap. 15C]{kay1993fundamentals}
\begin{equation}
\{\bJ_S\}_{i_h,j_q} = \tr \left ( \bSigma^{-1} \frac{\partial \bSigma}{\partial \t_{i_h}} \bSigma^{-1} \frac{\partial \bSigma}{\partial \t_{j_q}} \right ),
\end{equation}
where $i,j=1,\dotsc,\K$, and $h,q \in \{x,y\}$ index the two components of each 2D shift vector $\btau_i = [\tau_{i_x}, \tau_{i_y}]^T$.
Carrying out the indicated operations (see Appendix~\ref{app:crlbRand}) we get
\begin{equation}
%\bJ = \rho (\oV \oV^T - (\K + 1) \I),
\bJ_S = \left[ (\K + 1) \I_{\K} - \oV \oV^T \right] \otimes \mathbf{B},
\end{equation}
with 
\begin{equation}
\mathbf{B} = 
      \begin{bmatrix}
	\rho_{x,x} & -\rho_{x,y} \\      
        -\rho_{x,y}  & \rho_{y,y}
      \end{bmatrix},
\end{equation}
and
\begin{equation}
\rho_{h,q} = \sum_{l=1}^M  \frac{2S^2(\bomega_l)\, \omega_{l_h} \omega_{l_q}}{N^2(\bomega_l) + (K+1)S(\bomega_l)N(\bomega_l)}.
\label{eq:gammaHQ}
\end{equation}
Hence, we have
\begin{equation}
\bJ_S^{-1} = \tfrac{1}{(\K+1)} [ \I_K + \oV \oV^T ] \otimes \mathbf{B}^{-1}.
\end{equation}
The error covariance matrix of any unbiased  estimate of the shifts is thus bounded by
\begin{equation}
\E_{\bz|\btau}[(\hat{\btau} - \btau)(\hat{\btau} - \btau)^T] \geq \mathbf{J}_S^{-1},
\end{equation}
and the MSE is lower bounded by the trace of $\mathbf{J}_S^{-1}$,
\begin{align}
\mse &\geq \frac{1}{2\K}\tr(\bJ_S^{-1})  = \frac{1}{(\K+1)} \left ( \frac{\rho_{x,x}^2 + \rho_{y,y}^2 }{\rho_{x,x} \rho_{y,y} - \rho_{x,y}^2 } \right ).
\end{align}
If $S(\bomega)$ and $N(\bomega)$ are rotationally symmetric (i.e., rotation invariant), it can be shown that $\rho_{x,y}=0$ and $\rho_{x,x}=\rho_{y,y}$. In this case, we define the CRB under the Gaussian stochastic image model (CRBS) as
\begin{equation}
\crbs \mydef  \frac{2}{(\K+1) \rho_{xx}} = \frac{2}{(\K+1) \rho_{yy}}. %\left [ \frac{\rho_{x,x}^2 + \rho_{y,y}^2 }{\rho_{x,x} \rho_{y,y} } \right ].
\label{eq:crlbRand}
\end{equation}

Notice that, unlike the CRBD~\eqref{trBound}, the CRBS~\eqref{eq:crlbRand} depends on the number of images $\K$.

\smallskip

\noindent\textbf{High SNR.} Under high signal-to-noise conditions, the CRBS  for a rotation invariant process \eqref{eq:crlbRand} simplifies to
\begin{equation}
\crbs^\textsc{hsnr} \mydef  \frac{2\sigma^2 (2 \pi)^2}{ N_p \int S(\bomega) \omega^2_x \diff \bomega} . %\left [ \frac{\rho_{x,x}^2 + \rho_{y,y}^2 }{\rho_{x,x} \rho_{y,y} } \right ].
\label{eq:crbsHSNR}
\end{equation}
This bound is indeed independent of the number of images $K$ and agrees with the deterministic CRB given by~\eqref{trBound} (see Appendix~\ref{app:crlbRand}).

\smallskip

To help further understand the behavior of the computed bound we analyze its behavior both for natural and flat spectrum images.

\smallskip

\noindent \textbf{Natural images.} 
A typical natural image presents  complex structure that is difficult to model accurately. 
One classical assumption, it that the power spectrum of natural images falls quadratically with the Fourier frequency. 
Although simplistic, this is in fact reasonable if we consider that natural images have a relative contrast energy that is scale invariant~\cite{field1987relations}.
 Let us assume that the considered underlying image follows this law, that is,
\begin{equation}
S(\bomega ) = 
\begin{cases}
 S_n \|\bomega \|^{-2} & \text{if} \quad  \max(|\omega_x|,|\omega_y|)\le  W/2,\\
0 & \text{otherwise,}
\end{cases}
\label{eq:naturalSignal}
\end{equation}
where $\kn$ is a known parameter, and $W \in (0,2\pi]$ models the signal bandwidth. 
Also, we will assume that the additive noise spectrum has a constant value $N$ in the frequency band $[-\frac{W}{2}, \frac{W}{2}]^2$ and is zero otherwise.

Let us define the signal-to-noise ratio as the ratio between the energy of the derivative and the noise power
\begin{equation}
\snrg \mydef \frac{1}{N W^2}\int S(\bomega) \|\bomega\|^2 \diff \bomega.
\label{eq:SNR}
\end{equation}
For the case of a natural image~\eqref{eq:naturalSignal}, the \snrg{} is then
\begin{equation}
\snrg_n  = \kn/N.
\label{eq:SNRnat}
\end{equation}
and the CRBS bound for natural images becomes (see Appendix~\ref{app:crlbRand})
\begin{equation}
\crbsn \mydef \frac{8 \pi}{N_p (K+1)  \snr_n^2  \acoth \left( 1 \!+\! \frac{2\pi(K\!+\!1)\snr_n}{W^2} \right)}.
\label{eq:crbsn}
\end{equation}

When $\snrg_n \to \infty$, we have that
\begin{equation}
\crbsn \to \frac{16 \pi^2}{\Np W^2 \snrg_n},
\label{eq:limitN}
\end{equation}
which does not depend on $\K$. The breaking point from the asymptotic (very high SNR point) occurs approximately when
$2\pi(K\!+\!1)\snr_n \approx W^2$, which happens at,
\begin{equation}
\snrg^K_{n1} = \frac{W^2}{2\pi (K+1)}.
\label{eq:thCRBSN}
\end{equation}
This implies that, if $\snrg_n \gg \snrg^{K=1}_{n1}$, having access to more than two images will not improve the bound. 
This absolute breaking point happens at approximately  $\snrg_{n1} \mydef \frac{1}{5} \frac{W^2}{4\pi}$.

\medskip

\noindent \textbf{Flat spectrum images.}
Another helpful case is to study the behavior of the CRBS when the underlying signal $\u$ has a flat power spectral density, that is,
\begin{equation}
S(\bomega ) = 
\begin{cases}
S_w & \text{if }  \max(|\omega_x|,|\omega_y|)\le  W/2,\\
0 & \text{otherwise.}
\end{cases}
\label{eq:whiteSignal}
\end{equation}
Similarly, we assume that the additive noise spectrum $N(\bomega)$ has a constant value of $N$ in the frequency band $[-\frac{W}{2}, \frac{W}{2}]^2$ and is zero otherwise. 

The  signal-to-noise ratio (as defined in~\eqref{eq:SNR}) for white signals becomes
\begin{equation}
\snrg_w = \frac{\kw W^2}{6 N}.
\label{eq:SNRW}
\end{equation}
In this case, the CRB bound for flat spectrum images is (see Appendix~\ref{app:crlbRand})
\begin{align}
\crbsw & \mydef \frac{8 \pi^2 \left( W^2 + 6 \snrg_w(K+1) \right)}{3\Np(K+1)\snrg_w^2  W^2} \nonumber \\
&=  \frac{8 \pi^2}{3\Np(K+1)\snrg_w^2 } +  \frac{16 \pi^2   }{\Np W^2\snrg_w  }.
\label{eq:crbWhite}
\end{align}

In particular, when $\snrg_w \to \infty$ we have
\begin{equation}
\crbsw \to \frac{16 \pi^2   }{\Np W^2\snrg_w  },
\label{eq:limitW}
\end{equation}
which does not depend on the number of images $\K$. The breaking point is when both terms in~\eqref{eq:crbWhite} are approximately equal,
which happens  at
\begin{align}
\snrg^K_{w1} = \frac{W^2}{6 (K+1)}.
\label{eq:snr1White}
\end{align}
Thus, if $\snrg _w \gg  \snrg^{K=1}_{w1}$, having access to more than two images will not improve the bound.
This absolute breaking point happens at approximately  $\snrg_{w1} \mydef \frac{1}{5} \frac{W^2}{12}$. 

Because this threshold is very similar to the one obtained for natural images (see Eq.~\eqref{eq:thCRBSN}), for simplicity, we refer hereafter to both $\snrg_{n1}$ and $\snrg_{w1}$ as $\snr_1$.
\smallskip

\begin{figure}
\centering
\begin{minipage}[c]{.49\linewidth}
\centering
\includegraphics[width=.99\linewidth]{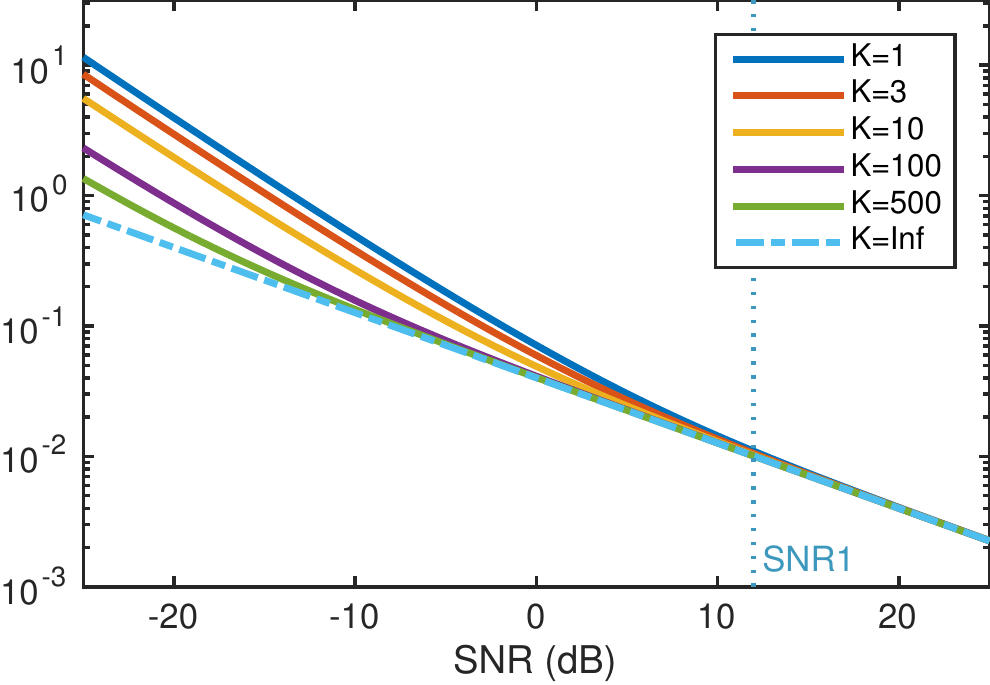}\vspace{-.2em}

{\scriptsize {(a)} $\textsc{crbs}_n$ (Eq.~\eqref{eq:crbsn})}
\end{minipage}
\begin{minipage}[c]{.49\linewidth}
\centering
\includegraphics[width=.99\linewidth]{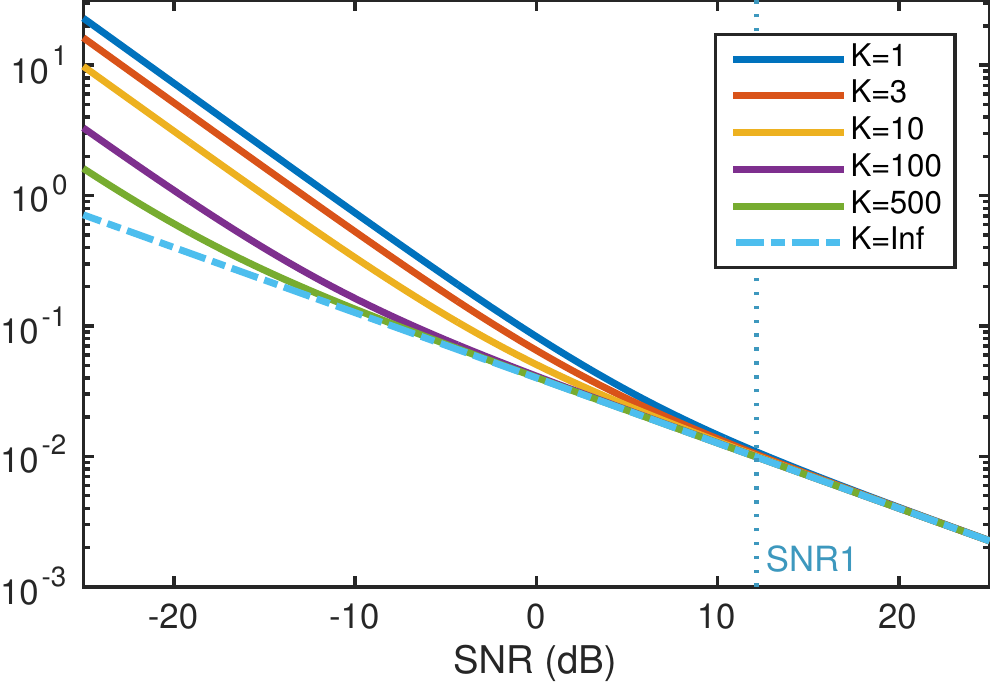}\vspace{-.2em}

{\scriptsize {(b)} $\textsc{crbs}_w$ (Eq.~\eqref{eq:crbWhite}}
\end{minipage}

\caption{Comparison of CRBS for different number of input images $K$ and varying SNR conditions. (a) natural image model, (b) flat spectrum image.
Both image models have very similar behavior.}
\label{fig:crbs_varM}
\end{figure}

Figure~\ref{fig:crbs_varM} shows the computed CRBS bounds for both image models, with different number of input images $K$ and varying SNR levels.
Both image models have very similar behavior. There is a very high SNR zone where the bounds depend linearly with the SNR level ($\textsc{snr}>\textsc{snr}_1$).
Within this SNR region, having access to more images does not have an impact on the bound. In moderate to low SNRs ($\textsc{snr}<\textsc{snr}_1$)
both Cram\'er-Rao stochastic bounds depend super-linearly with the SNR (i.e., performance degrades faster at low SNR values than in the very high SNR region). 
Increasing the number of images pushes back $\snr_1$, increasing the SNR range where performance is linear with SNR. The performance is linear with image size $N_p$ in both cases on 
the whole SNR domain.

\smallskip

An alternative approach to include an image model is to compute a Hybrid Cram\'er-Rao bound (HCRB).\footnote{\emph{Hybrid} in the sense that there are random and deterministic parameters.}
Similarly to what was done in Section~\ref{ssec:BCRBshifts}, one can compute a HCRB including 
the desired model as an image prior and then compute the expected FIM under this prior. 
Robinson and Milanfar~\cite{robinson2006statistical} computed such bound
for the super-resolution problem assuming a Gaussian model for the image similar to the one presented
here. 
Although related, these two bounds are different. The HCRB gives a bound on the expected MSE under the given image prior. 
This can be seen as an average bound for the different likelihoods obtained for each given possible value of the image. 
On the other hand, the CRB on Equation~\eqref{eq:crlbRand} gives the bound based on the expected likelihood under the given image model. 
Under some regularity conditions, it is possible to show that the CRB is always tighter than the HCRB~\cite[Thm.~1]{noam09notes}. 
Nevertheless, in many applications the computation of the HCRB is much simpler than the CRB, leading to a reasonable alternative.

\subsection{Extended Ziv-Zakai lower bound}
In general, the CRB is known to be tight in high SNR but overoptimistic in low SNR conditions. Various Bayesian bounds have been derived to obtain tighter and more accurate predictions of the MSE behavior in the entire SNR range. One example of this is the bound proposed by Ziv and Zakai~\cite{ziv1969some}, which relates the expected MSE (EMSE) of the estimator over a given prior, to the probability of error in a binary detection problem. 

Consider the estimation of a $2K$-dimensional random vector $\btheta$ with a prior distribution $p_{\btheta}$, based upon an observation vector $\bz$.
The extended Ziv-Zakai lower bound (EZZB) on the EMSE of any estimate $\hat{\btheta}$ of $\btheta$ over $p_{\btheta}$ is given by \cite{bell1997extended}
\begin{align}
\ba^T  \overbar{\bR}_\epsilon \ba \ge \int_0^\infty \nuC \left\{ \max_{\bdelta:\ba^T \bdelta = h} \right. &\bigg[ \int_{\Rk} \min (p_{\btheta}(\bphi),p_{\btheta}(\bphi + \bdelta))  \nonumber \\
															               &\!\!\!\!\!\!\!\left.  \cdot P_\text{min} (\bphi,\bphi + \bdelta) \diff \bphi \bigg] \right\} h \diff h,
\label{eq:ezzb}
\end{align}
where $\ba$ is any $2K$-dimensional vector, $\nuC\{\cdot\}$ is the valley-filling function,%
\footnote{The valley-filling of a function $f(h)$ is obtained by filling-in any valleys~\cite{bell1997extended}, and is given by $\nuC\{f\}(h) = \max_{t\ge0} f(h + t)$.}
and $P_\text{min} (\bdelta)$, $\bdelta \in \mathbb{R}^{2K}$, is the probability of error in the binary detection problem 
\begin{align}
&H_0: \hat{\bdelta} = \bphi;  \bz \sim  p(\bz \cond \btheta = \bphi),\\
&H_1: \hat{\bdelta} = \bphi + \bdelta; \bz \sim  p(\bz \cond \btheta = \bphi + \bdelta),
\label{eq:ezzbHyp}
\end{align}
with equally likely hypotheses. The vector $\bdelta = [\bdelta_1,\ldots,\bdelta_K]$, with $\bdelta_i = [\delta_{i_x}, \delta_{i_y}]^T$ represents a possible 2D shift between the $i$-th and the first image (indexed in the same way as $\btau$).

The Ziv-Zakai bound is based on the probability of correctly choosing the parameter to be estimated between two possible values:  $\bphi$ or $\bphi + \bdelta$. 
The bound is found by integrating the minimum error along all possible estimated values (in general ruled by both $\bdelta$ and $\bphi$), weighted by their prior probability of occurrence, and by bounding the minimum probability of error in this binary detection problem.

If the probability of error is only a function of the offset between the hypothesis, i.e., $P_\text{min} (\bphi,\bphi + \bdelta) = P_\text{min}(\bdelta)$, which is precisely the case in our translation estimation problem, the bound simplifies to
\begin{equation}
\ba^T \overbar{\bR}_\epsilon \ba \ge \int_0^\infty \nuC \left\{ \max_{\bdelta:\ba^T \bdelta = h} A(\bdelta) P_\text{min} (\bdelta) \right\} h \diff h,
\label{eq:ezzbSimpl}
\end{equation}
where
\begin{equation}
A(\bdelta) = \int_{\mathbb{R}^{2\!K}}  \min \left( p_\btheta(\bphi),  p_\btheta(\bphi + \bdelta) \right) \diff \bphi.
\label{eq:ezzbA}
\end{equation}
Thus, to compute the EZZB of the shift estimation problem we need to compute $A(\bdelta)$ and the probability of error $P_\text{min}(\bdelta)$.

If we assume the shifts $\btheta$ to be uniformly distributed $\btheta \sim \mathcal{U}[0,D]^{2K}$, $A(\bdelta)$ takes the simplified form
\begin{equation}
A(\bdelta) = \prod_{i=1}^{2K} \Big(1 - \frac{\delta_i}{D}\Big).
\label{eq:A}
\end{equation}
The probability of error $P_\text{min} (\bdelta)$ for the case of multi-image registration is given by (see Appendix~\ref{sec:probErrorEZZB})
\begin{align}
P_\text{min} (\bdelta)  \approx \tfrac 12 \exp \left\{ a(\bdelta) + b(\bdelta) \right\} \Phi \left(\sqrt{ 2 b(\bdelta)}\right),
 \label{eq:pe}
\end{align}
where 
\begin{equation}
a(\bdelta) \!=\!  -\sum_{l=1}^M \log \big [ 1 \!+\! \gamma(\bdelta, \bomega_l) \big ],\,\, b(\bdelta) \!=\!    \sum_{l=1}^M \frac{\gamma(\bdelta, \bomega_l)}{1\!+\!\gamma(\bdelta, \bomega_l)},\label{eq:Peb}
\end{equation}
\begin{equation}
\gamma(\omega,\bdelta) = \frac{ S(\bomega)^2  \Big( (K+1)^2 - T(\bdelta,\bomega) \Big)}{4\left( N(\bomega)^2 + (K+1)N(\bomega)S(\bomega)\right)},
\end{equation}
\begin{equation}
T(\bdelta,\bomega) =\Big\lvert1 + \sum_{j=1}^K e^{-i \bdelta_j \cdot \bomega}\Big\rvert^2 \,\, \text{and} \,\, \Phi(t) = \tfrac{1}{\sqrt{2 \pi}}\int_t^\infty e^{-\frac{t^2}{2}} \diff t.
\end{equation}

\smallskip

\noindent \textbf{Flat spectrum signals.}
As done for the CRBS case, let us consider the particular case of flat spectrum signals defined previously by Equation~\eqref{eq:whiteSignal}. The analysis presented hereafter closely follows and extends the work by Weinstein and Weiss~\cite{weinstein1984fundamental} to the case of multiple signals.

For simplicity, the following analysis is restricted to one-dimensional signals. 
The  extension to two-dimensional signals is straightforward in the case where the image is assumed to be drawn from a white random process (full bandwidth flat spectrum, i.e., $W=2\pi$) . 
In this case, knowing the translation in one direction does not give any additional information to the estimation of the other one. 
As a consequence, the 2D image can be rearranged into a one-dimensional vector by concatenating its rows without loss of information regarding the estimation of the
translation along the columns.
Following this remark, in this section, we will consider one-dimensional signals having length $\Np=m_r \times m_c$ and $W=2\pi$.

\smallskip

The EZZB corresponding to the estimation of one single component is given by (see Appendix~\ref{app:ezzbWS})
\begin{align}
\msec   \ge       \ezzbw& \mydef\tfrac{1}{c^2} \int_0^{\sqrt{2b}} h \exp \left\{-\tfrac{9 h^4 }{20 \Np}\right\} \Phi(h) \diff h \nonumber \\
          &\quad\phantom{\mydef}+ \tfrac{D^2}{6} e^{a+b} \Phi(\sqrt{2b}),
\label{eq:ezzbWS}
\end{align}
where
\begin{align}
&a =\! -N_p \log \left( \tfrac{ \sqrt{\kappa_2+1}+1}{2}\right),  b = \tfrac{N_p}{2} \tfrac{ \sqrt{\kappa_2+1} -1}{\sqrt{\kappa_2+1}},  c^2\! = \tfrac{ N_p \pi^2 \kappa_1}{ 12}\nonumber\\
&\kappa_1 = \tfrac{ 9 \snrg_w^2 (K+1)}{8\pi^4 + 12\pi^2 \snrg_w(K+1)}, \,\, \kappa_2 =  \tfrac{ 9 \snrg_w^2 K}{4\pi^4 + 6\pi^2 \snrg_w(K+1)}.
\end{align}

\medskip

\begin{figure}
\centering
\begin{minipage}[c]{.49\linewidth}
\centering
\includegraphics[width=\linewidth]{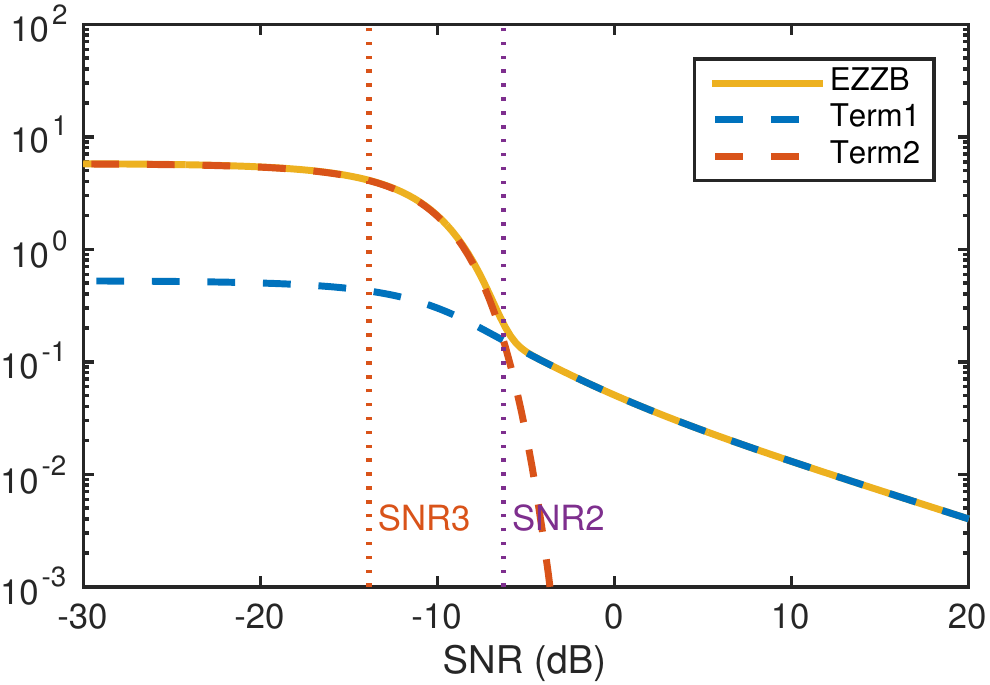}\vspace{-.3em}

{\scriptsize (a)}
\end{minipage}
\begin{minipage}[c]{.49\linewidth}
\centering
\includegraphics[width=\linewidth]{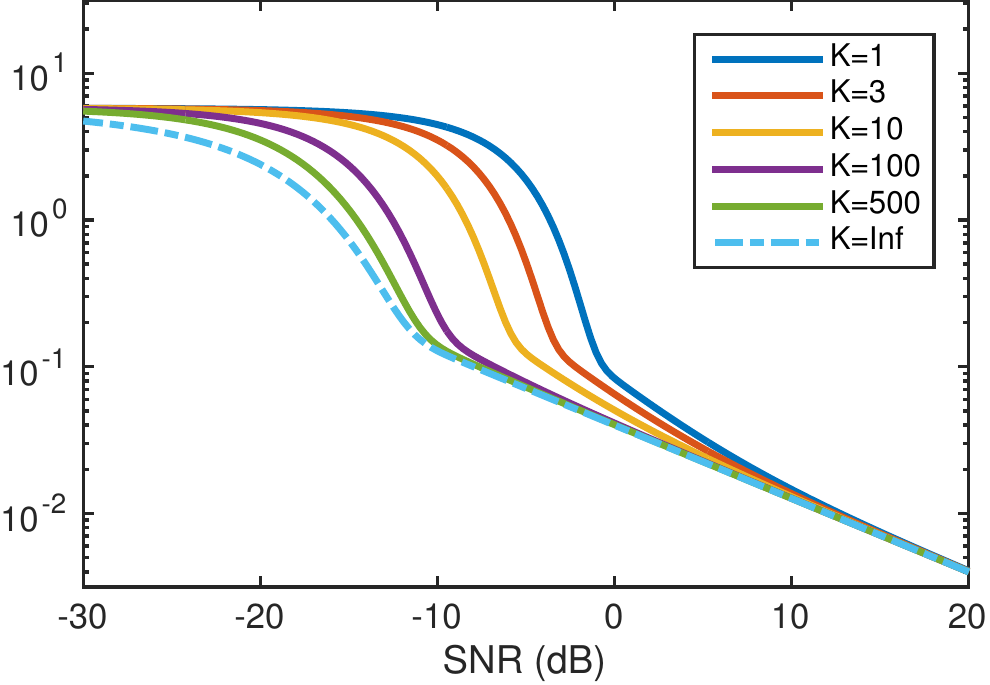}\vspace{-.3em}

{\scriptsize (b)}
\end{minipage}
\vspace{.7em}

\begin{minipage}[c]{.49\linewidth}
\centering
\includegraphics[width=\linewidth]{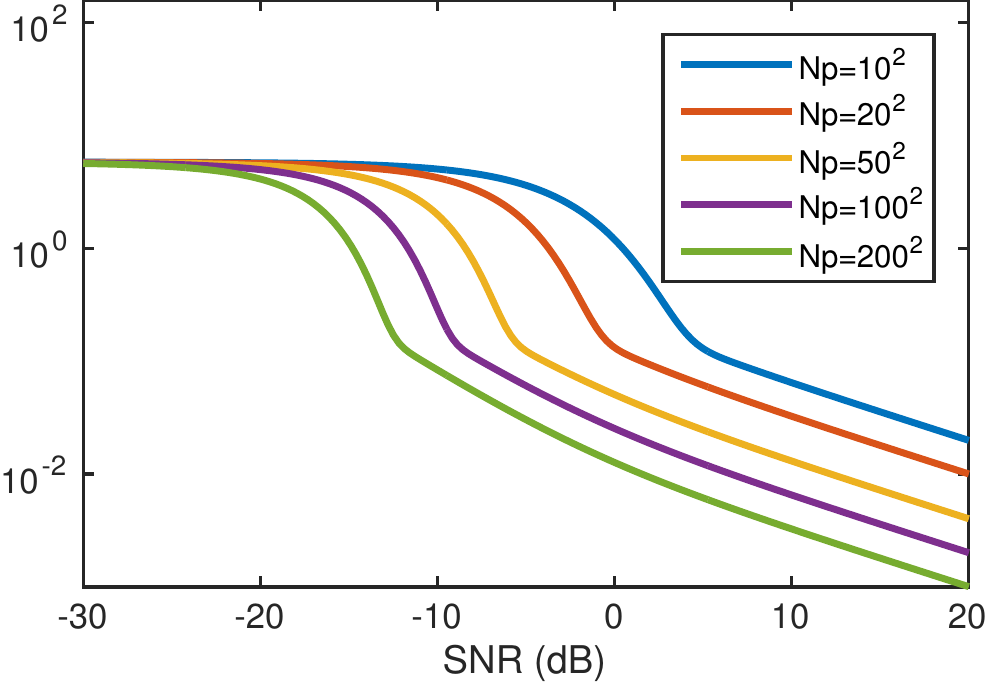}

{\scriptsize (c)}
\end{minipage}
\begin{minipage}[c]{.49\linewidth}
\centering
\includegraphics[width=\linewidth]{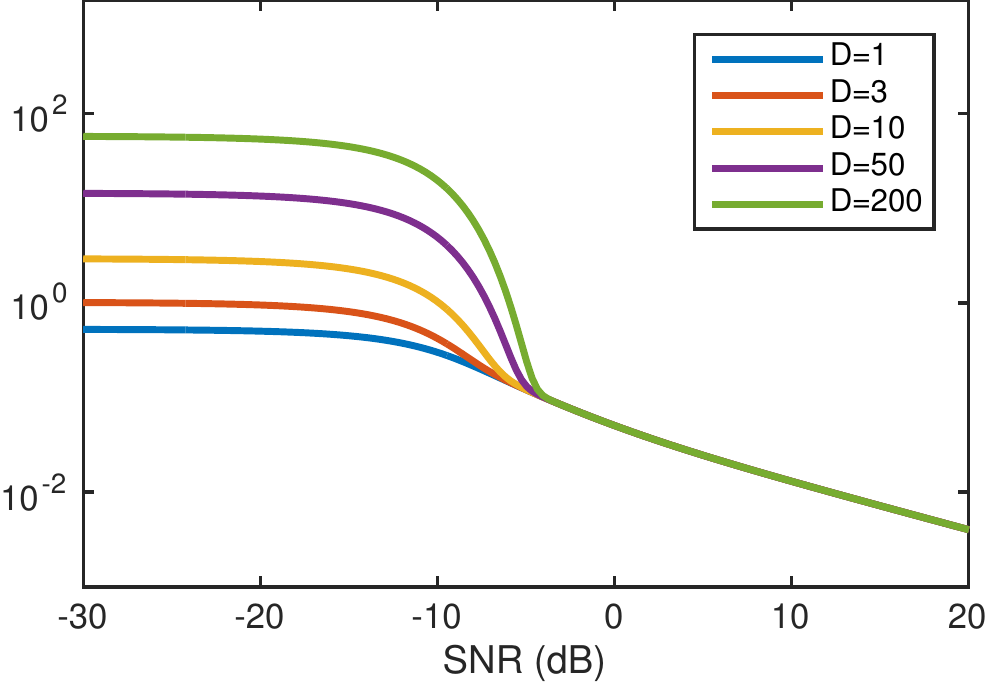}

{\scriptsize (d)}
\end{minipage}

\caption{(a) The breaking point of the EZZB bound and its decomposition as the sum of two terms for varying SNR conditions and $K=10$. 
Term 1 corresponds to Equation~\eqref{eq:EZZBterm1} and term 2 corresponds to Equation~\eqref{eq:EZZBterm2}. 
(b-d) Comparison of EZZB at varying SNR conditions for different number of input images $\K$ (b), different image size $N_p$ (c) and different shift prior intervals $D$ (d).
}
\label{fig:ezzbCompare}
\end{figure}

\noindent \textbf{Analysis of the EZZB: different SNR regions.}
The EZZB behaves differently in low and high SNR regimes, as dictated by the two terms in Eq.~\eqref{eq:ezzbWS} and
as illustrated in Figure~\ref{fig:ezzbCompare}(a). 

\medskip

\noindent \emph{i) High SNR}. For $\snr_w \gg 1$, the error term~\eqref{eq:ezzbWS} is mainly driven by the first term since $a+b \to -\infty$ and 
$b \to \frac{\Np}{2}$.

Assuming that $\Np\gg1$, we obtain~\cite{weiss1983fundamental}
\begin{align}
\ezzbw &\xrightarrow{\snr_w \rightarrow \infty}  \tfrac{1}{c^2} \int_0^{\sqrt{2b}} h \cdot \exp \left\{-\tfrac{9 h^4 }{20 \Np}\right\} \Phi(h) \diff h \label{eq:EZZBterm1}\\
&\approx \tfrac{1}{c^2} \int_0^{\infty} h \cdot \exp \left\{-\tfrac{9 h^4 }{20 \Np}\right\} \Phi(h) \diff h = \frac{1}{4 c^2}\\
%&= \frac{1}{4 c^2} = \frac{6(N^2+NS_w(K+1)}{\pi^2 \Np S_w^2(K+1)} = \crbsw.
&= \frac{8\pi^2 + 12 \snrg_w(K+1)}{3 N_p \snrg_w^2(K+1)} = \crbsw.
\end{align}

Indeed, in the high SNR regime, the EZZB approaches the CRBS under the stochastic image model given by \eqref{eq:crbWhite}.

\smallskip

\noindent \emph{ii) Low SNR}. On the other hand, in a very low SNR scenario, $\snr_w \ll 1$, $a \to 0$, $b\to 0$ and $c\to 0$. Thus,
\begin{equation}
\ezzbw \xrightarrow{\snr_w \rightarrow 0}   \tfrac{D^2}{6} e^{a+b} \Phi(\sqrt{2b}) \approx \tfrac{D^2}{12} \label{eq:EZZBterm2},
\end{equation}
which is the variance of the shifts prior.

\smallskip

\noindent \emph{iii) Transition zone}. The transition from the low-SNR to the high-SNR region starts when the two terms have a similar contribution to the bound, that is, 
\begin{equation}
\frac{1}{4c^2} = \frac{D^2}{6}e^{a+b} \Phi(\sqrt{2b}).
\label{eq:SNR2}
\end{equation}
We could (arbitrarily) say that the transition is completed when the bound reaches half the asymptotic value, i.e.,
\begin{equation}
e^{a+b}\Phi(\sqrt{2b}) = \tfrac{1}{4}.
\label{eq:SNR3}
\end{equation}
Equations \eqref{eq:SNR2} and \eqref{eq:SNR3} characterize the limit SNR levels of the transition zone. 
Let $\snr_2$ be the SNR level that satisfies~\eqref{eq:SNR2} and $\snr_3$ the SNR level satisfying~\eqref{eq:SNR3}. 
Within this transition region the bound is essentially dominated by the behavior of $\Phi(\sqrt{2b})$.

Figure~\ref{fig:ezzbCompare} shows how this region changes when varying the number of images $K$, the prior  $D$ and the image size $N_p$.
The threshold $\snr_2$, below which the EMSE decreases significantly and worsens exponentially with the SNR, 
depends on the number of available images $\K$. 
This is probably the most important consequence of having access to more images. 
Figure~\ref{fig:ezzbCompare}(b) shows how this threshold can be pushed back several dBs by increasing $K$, until reaching a limit.

Note that this critical SNR level $\snr_2$ also depends on the image size $N_p$. As a consequence, increasing the image size reduces the performance bound and 
pushes this SNR limit as shown in Figure~\ref{fig:ezzbCompare}(c).

On the other hand, as illustrated in Figure~\ref{fig:ezzbCompare}(d), the threshold $\snr_2$ is not significantly affected by the shift 
prior parameter $D$. This means that, for practical $D$ values (e.g., $D \geq 1$), having a tighter shift prior does not push 
back $\snr_2$ significantly. Nevertheless, as expected, the steady state EMSE predicted by EZZB does decrease with $D$.

\section{Comparison of performance bounds}
\label{sec:boundsAnalysis} 
In this section, we analyze and compare the behavior of the previously computed CRB and EZZB bounds. To this effect, it is important to make the distintion that the CRB is a bound on the MSE while the EZZB and the BCRB are bounds on the EMSE over a given prior for the shifts. 

\begin{figure}
\centering
\includegraphics[width=.7\linewidth]{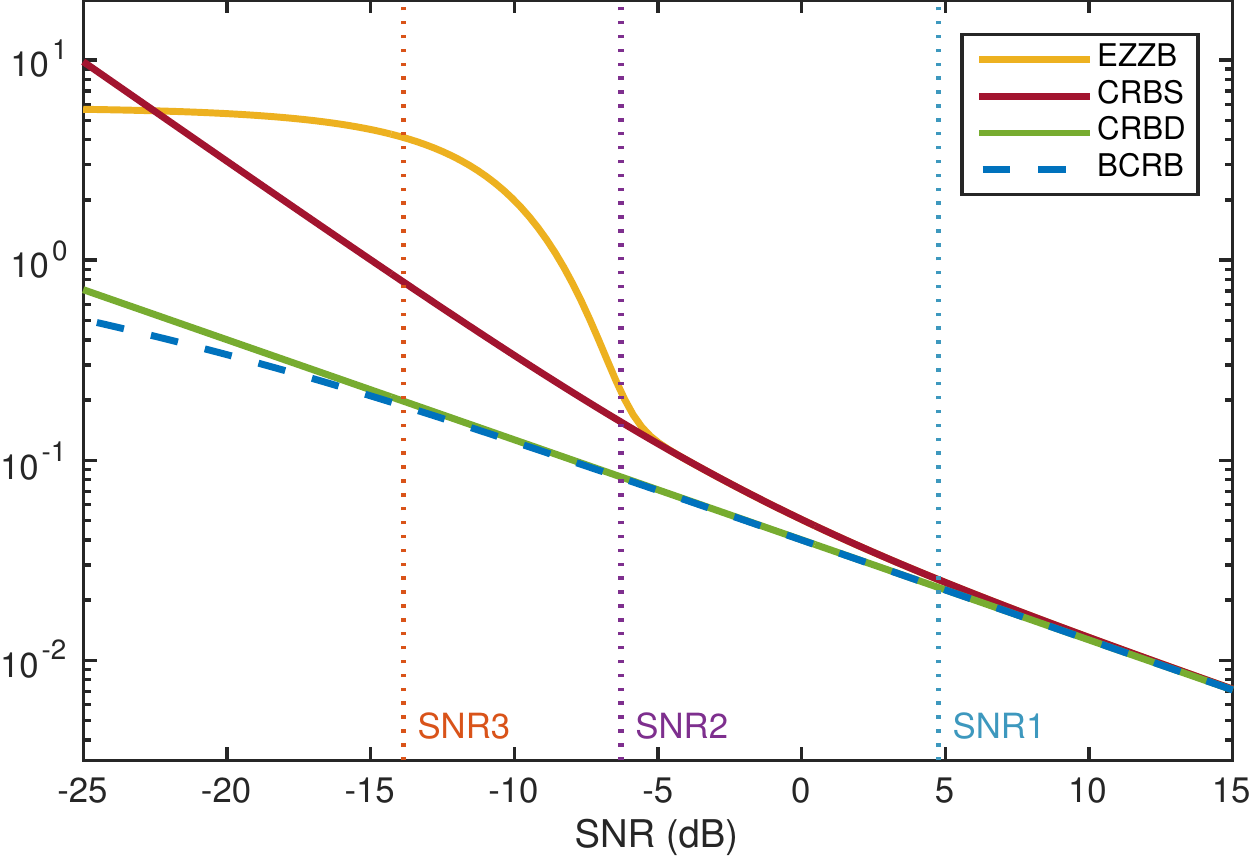}
\caption{Comparison of the EZZB, CRBS, CRBD and BCRB bounds for $K=10$ and varying SNR conditions.}
\label{fig:comparBoundsK1}
\end{figure}

To simplify the discussion, let us consider the case of white signals. Figure~\ref{fig:comparBoundsK1} shows a comparison of the CRB bounds (both for deterministic (CRBD) and stochastic (CRBS) image models), the BCRB (with Gaussian shift prior of variance $\lambda^2=1$) and the EZZB (with uniform $[0,D]$ shift prior, $D=20$) assuming an image of size $50\times50$ pixels. For the CRBD and the BCRB cases, that depend on a deterministic signal, we used a realization from the  white random process used in CRBS and EZZB. Based on the SNR values, the behavior of the bounds can be characterized into four different regions i-iv.
\smallskip

\noindent \textbf{i) Very high SNR} ($\snr \ge \snr_1$). In this region, all bounds agree. Hence, the same fundamental limit is predicted for both the MSE and the EMSE. This limit does not depend on the shift value nor on the width of the prior, within practical limits for $\lambda$ and $D$ ($D,\lambda \geq 1$). 
The performance bound only depends on the total image gradient energy and the noise level, and it is linear with the SNR and image size $N_p$. Hence, a very important remark is that, in this SNR region, all bounds predict that having access to more than two images ($K>1$) or having a more accurate shift priors (a smaller $\lambda$ or $D$ within practical limits) will not lead to better performance. 
The threshold defining this region, $\snr_1$, depends on the number of images $\K$ (see Eq.~\eqref{eq:snr1White}). It does not depend, however, on the variance of the prior ($D$,$\lambda$) nor the image size ($N_p$). 

\smallskip

\noindent \textbf{ii) High SNR} ($\snr_1 \le \snr \le \snr_2$). In this region, the CRBS and the EZZB agree, while the CRBD and the BCRB are overoptimistic. The main differences with respect to the very high \snr{} region is that the CRBS and the EZZB improve with increasing number of images $\K$, and their dependence on the SNR is super-linear. This means that the performance decreases faster when reducing the SNR than in the very high SNR region.  In the limit, when $K\to \infty$, the EZZB and CRBS approach the CRBD (same behavior as in very high SNR). In this region, performance is linear with image size $N_p$.

\smallskip
  
\noindent \textbf{iii) Transition} ($\snr_2 \le \snr \le \snr_3$). The EZZB predicts a threshold $\snr_2$ below which the EMSE decreases significantly, and worsens exponentially with the SNR. This critical SNR level can be improved by increasing the number of available images $\K$ (up to some limit, see Figure~\ref{fig:ezzbCompare}(b)) or the image size $N_p$. Nevertheless, the threshold does not depend considerably on the shift prior $D$ (see Figure~\ref{fig:ezzbCompare}(d)). 

\smallskip

\noindent \textbf{iv) Saturation} ($\snr \le \snr_3$).
The EZZB predicts a critical SNR below which no alignment is possible, and thus the error is dominated by the shifts prior (the EMSE is essentially given by the variance of the prior).

\section{Performance bounds tightness assessment}
\label{sec:mle}
The bounds derived in sections~\ref{sec:detImage} and~\ref{sec:randImage} set an upper limit on the best possible performance of any estimator, but there is no guarantee about the existence of an estimator reaching that performance. 
Therefore, by only looking at the bounds, it is hard to draw practical conclusions about the actual achievable alignment performance in practice. 
Indeed, there could always exist a tighter bound, with a different behavior than the computed ones, that gets closer to best achievable performance.
Hence, assessing the tightness of the derived bounds to the actual alignment performance becomes critical to close this gap. 

In what follows, we compare the empirical performance of the maximum likelihood estimator (MLE) to the bounds previously computed. 
MLE is perhaps the most widely used estimator in statistical parameter estimation problems. 
It is asymptotically efficient~\cite{kay1993fundamentals}, and it is also known 
to be efficient for any number of samples in various problems~\cite{aguerrebere14}. 

\subsection{Maximum Likelihood Estimation}
Given $(\K+1)$ independent samples following Model~\eqref{eq:model}, and assuming $\bu$ 
is an unknown deterministic image, the MLE of $\btheta=[\bu, \btau]^T$ is the value that maximizes the log-likelihood,
\begin{equation}
[\bu, \btau]_\textsc{mle} =  \argmax_{\bu,\btau} - \frac{1}{2\sigma^2}\sum_{i=0}^{\K}||\z_i(\bx) - \u(\bx - \btau_i)||^2,
\label{mleFunc}
\end{equation}
where we discarded the terms independent of $[\bu, \btau]$.

The functional in~\eqref{mleFunc} is an example of a separable non-linear least-square problem. 
Indeed, given the vector $\btau$ containing all the shifts, the unknown underlying image $\bu$ would be given by the least squares solution
\begin{equation}
\hat{u}(\bx) = \frac{1}{(\K+1)} \sum_{i=0}^{\K}  z_i(\bx + \btau_i).
\label{mleU}
\end{equation}
That is, given the shift values, the MLE of the unknown image is the average of the aligned frames. 
Inserting~\eqref{mleU} back into~\eqref{mleFunc}, the functional to be optimized depends on the shifts only, that is,
\begin{equation}
\btau_{\textsc{mle}}  = \argmin_{\btau}  \sum_{i=1}^{\K} ||z_i(\bx)  -  \hat{u}(\bx-\btau_i)||^2,
\label{mleComb}
\end{equation}
where $\hat{u}(\bx)$ is given by \eqref{mleU}.
Functional~\eqref{mleComb} is non-convex and different approaches can be followed to find a local minimum~\cite{robinson2009optimal}. One such approach consists in alternating two steps: first compute the average of the frames aligned with the current estimate of the shifts (given by~\eqref{mleU}); second align each image against the current average by choosing the shift that maximizes the Euclidean distance against the average. That is,
\begin{align}
&\hat{u}^{(t+1)}(\bx) = \frac{1}{(\K+1)} \sum_{i=0}^{\K} z_i(\bx+\btau^{(t)}_i),\\
&\btau^{(t+1)}_i = \argmin_{\btau_i} || \bz_i - \hat{u}^{(t+1)}(\bx-\btau_i)||^2.
\end{align}
This algorithm requires an initialization either for $\btau$ or $\hat{u}$. One possibility is to align each input image to a reference image in the set and take those estimated translations as initial values. 
The algorithm stops when the shifts reach a steady value. 

In our implementation, we used image correlation~\cite{guizar2008efficient} which can be seen as an approximation of the $L_2$ distance that should be minimized. We initialize the iterative algorithm by aligning each image to the first one in the set,  and take those translations as initial values. We refer hereafter to this approximation of the MLE as \mleA{}. 

\subsection{Experimental analysis} 
\begin{figure}
\centering
\includegraphics[width=0.24\linewidth]{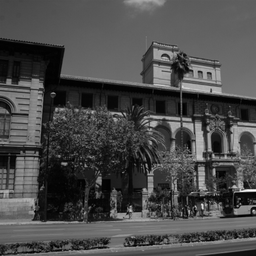}
\includegraphics[width=0.24\linewidth]{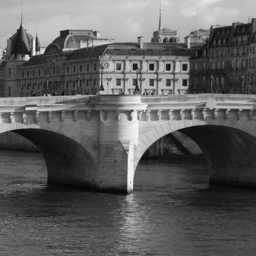}
\includegraphics[width=0.24\linewidth]{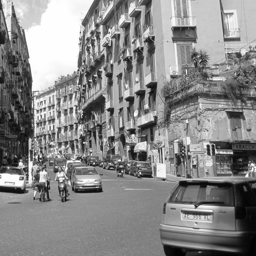}
\includegraphics[width=0.24\linewidth]{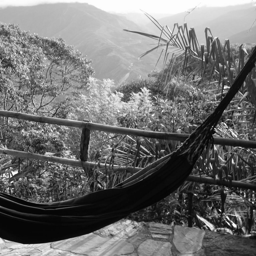}
\caption{Natural images used for the experimental analysis in Section~\ref{sec:mle}. From left to right: \texttt{building} (by M. Colom / CC BY), \texttt{paris}, \texttt{napoli}, \texttt{bolivia}. All images are $256\times256$ pixels.}
\label{fig:gtruthMLE}
\end{figure}

An experimental analysis is conducted in order to compare the performance of the
MLE to the previously introduced bounds. For this purpose, synthetic data is generated
according to model~\eqref{eq:model}. Two cases are considered for the underlying
image $\bu$: a natural image (Figure~\ref{fig:gtruthMLE}) and a flat spectrum
image (a realization of a uniformly distributed random variable). 
The shifts $\btau_i$ are uniformly sampled in  $[-5,5]^2$. 
Different noise levels $\sigma^2$ and number of images $K$ are evaluated. 
The performance was computed by averaging the estimation errors over 100 tests for each particular configuration ($K,\sigma^2$).

The squared bias of the \mleA{} experiments was on average, in all the conducted experiments, four orders of magnitude smaller than the estimator variance. Hence, we report the root mean squared error (RMSE) of the \mleA{}, which is dominated by its variance given that the method is almost unbiased. %\begin{figure}

\begin{figure}
\centering
\includegraphics[width=.4925\linewidth]{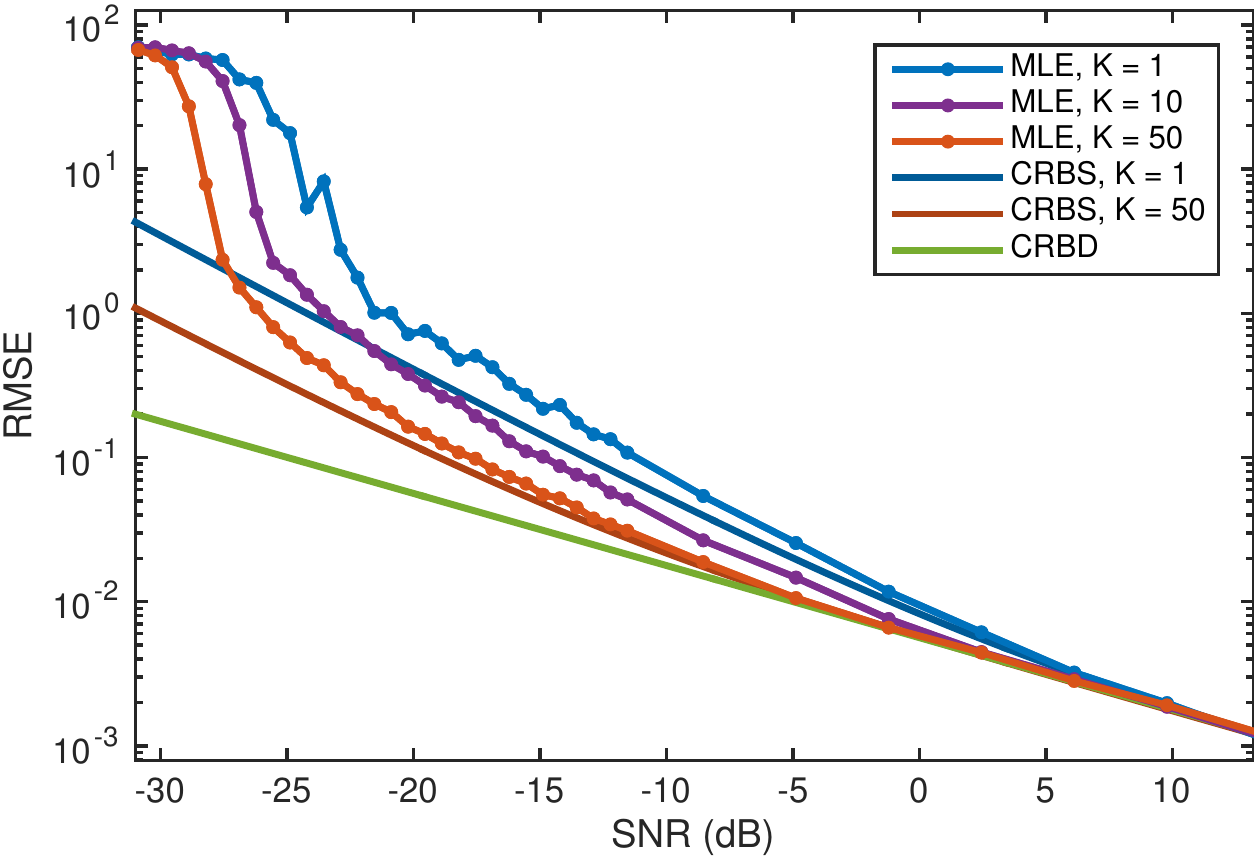} \includegraphics[width=.4925\linewidth]{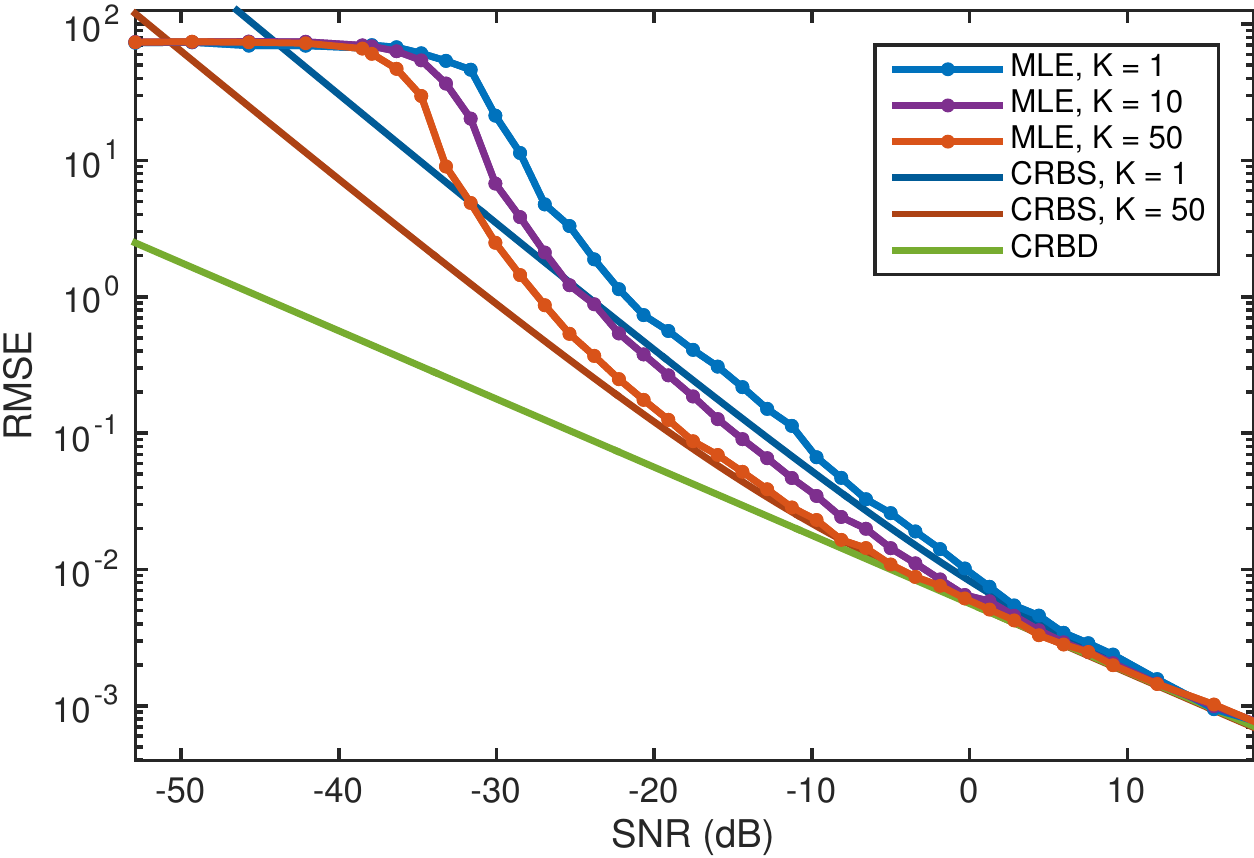}\\
\includegraphics[width=.4925\linewidth]{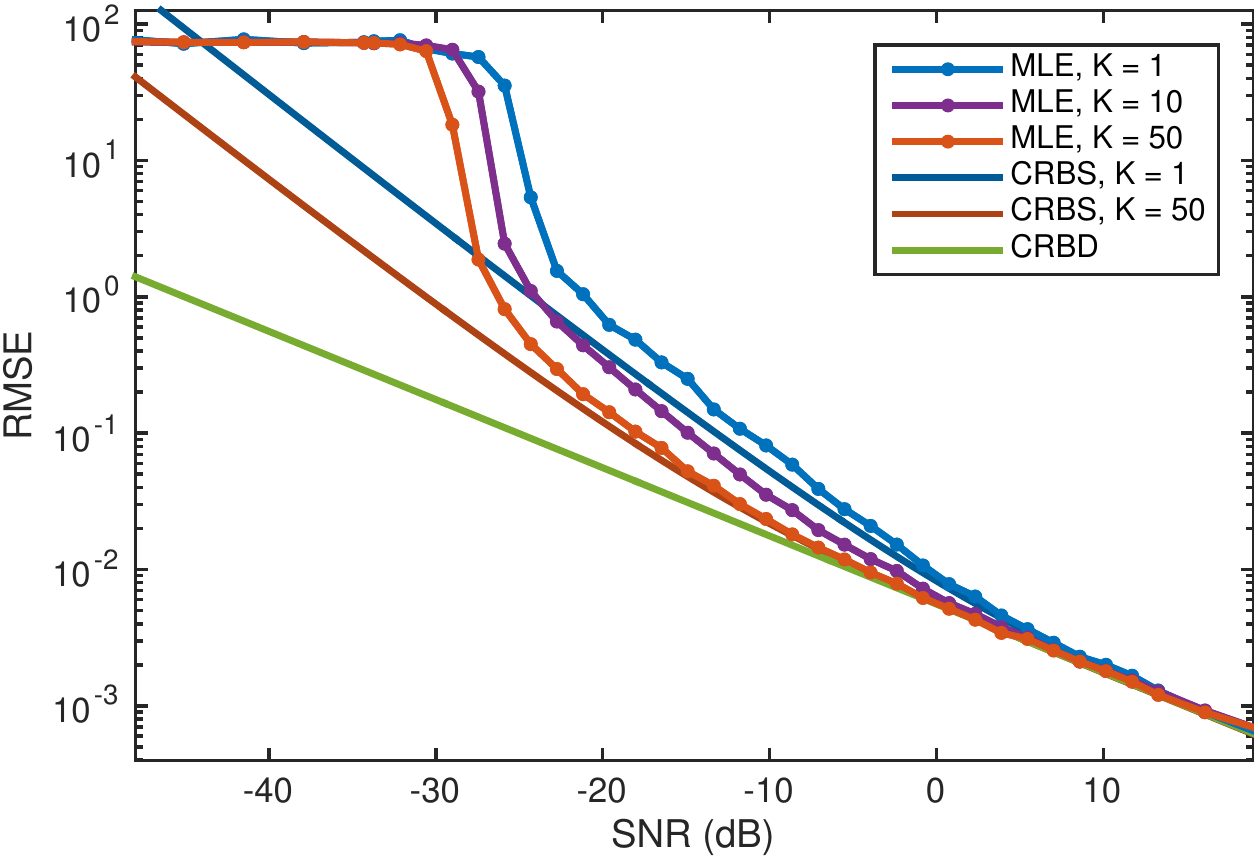} \includegraphics[width=.4925\linewidth]{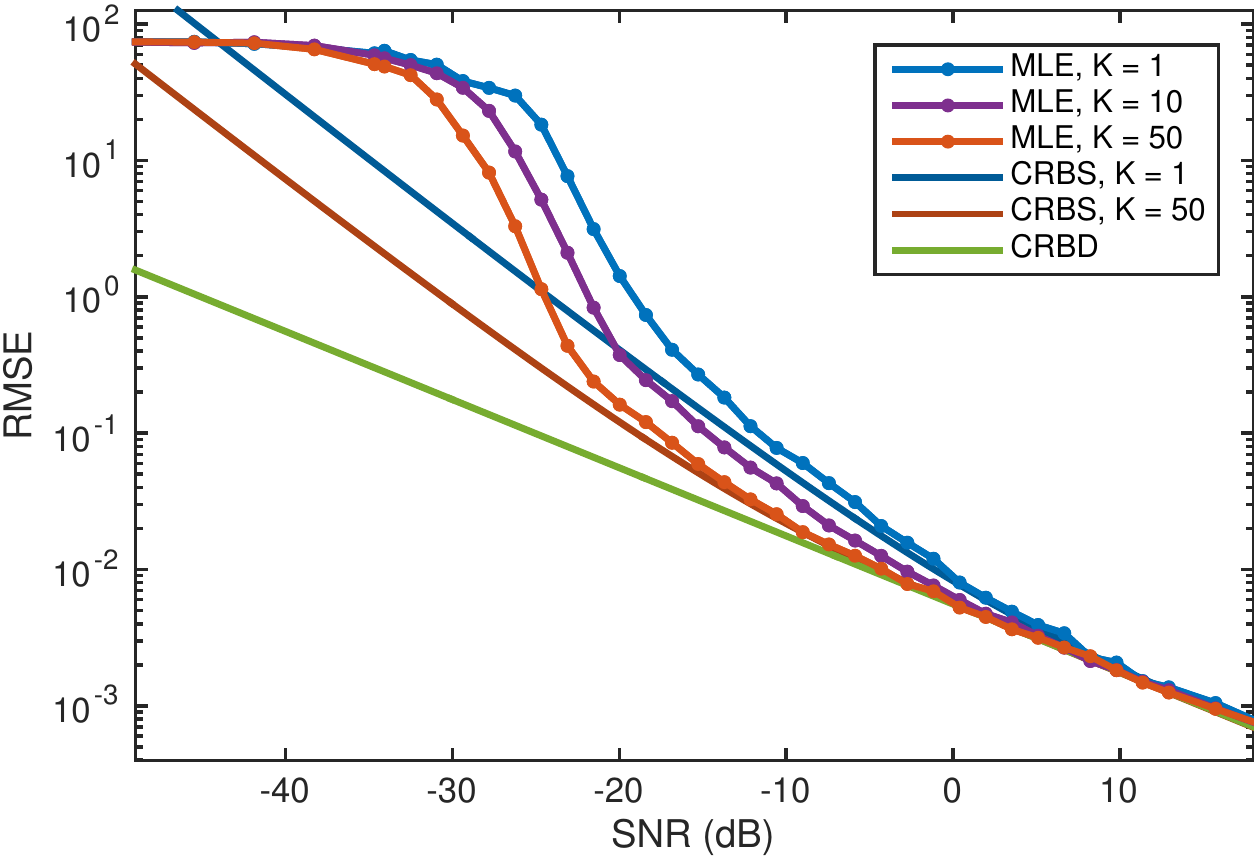}
\caption{\textbf{Natural images.} Comparison of the MLE performance for different number of images $\K$ to the CRB under the deterministic image model (CRBD) and the natural image stochastic model (CRBS) for the four examples shown in Fig.~\ref{fig:gtruthMLE}:  \texttt{building} (top-left), \texttt{paris} (top-right), \texttt{napoli} (bottom-left) and \texttt{bolivia} (bottom-right). }
\label{fig:comparMLEK}
\end{figure}

\begin{figure}
\centering
\includegraphics[width=.7\linewidth]{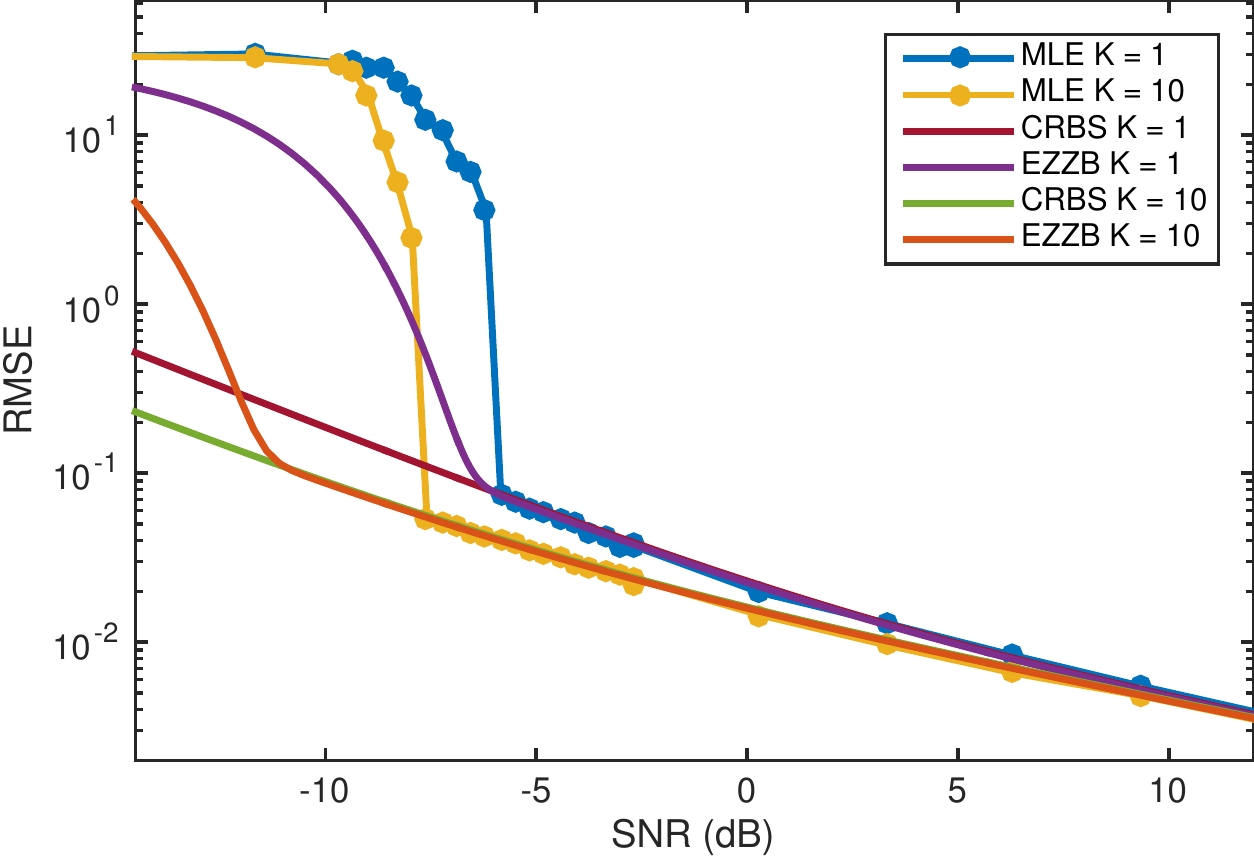}
\caption{\textbf{Flat spectrum.} Comparison of the MLE performance to the Cram\'er-Rao bound (CRBS) and the Ziv-Zakai bound (EZZB) for the pairwise and $\K=10$ cases.}
\label{fig:comparMLEBoundsNoiseK}
\end{figure}

Figure~\ref{fig:comparMLEK} shows the results obtained by the \mleA{} using as underlying image the ones shown in Figure~\ref{fig:gtruthMLE}, for different number of input images ($K=1,10,50$) and SNR levels. The results are compared to the CRBD (Eq.~\eqref{trBound}) and to the CRBS (Eq.~\eqref{eq:crbsn}) for $\K=10$. Figure~\ref{fig:comparMLEBoundsNoiseK} shows the results of the \mleA{} for $K=1,10$ on a flat spectrum underlying image, compared to the CRBS and EZZB. The case $\K=1$ corresponds to pairwise alignment.  

For both natural and flat spectrum images, similarly to what was predicted by the EZZB, we identify four different regions of behavior of the \mleA{} depending on the SNR value (compare figures~\ref{fig:comparMLEK} and~\ref{fig:comparMLEBoundsNoiseK} to Figure~\ref{fig:comparBoundsK1}). 

For very high SNR, all bounds agree and  the \mleA{} attains the limiting performance, which is independent of the number of images $\K$. Thus, under very high SNR, the alignment can be performed
pairwise without loss of accuracy and MLE is an optimal estimator.

For moderate to high SNR, a different behavior is observed for flat spectrum and natural images. For flat spectrum images, the \mleA{} still attains the limiting performance given by the CRBS and is thus optimal. For natural images, on the contrary, the \mleA{} performance is close to the CRBS but it is not tight. A possible reason explaining this behavior is the non-optimality of the \mleA{}, for which a critical drawback is that it does not use image prior information. Moreover, for this SNR region, the \mleA{} performance clearly improves with increasing number of images. Because the CRBS bound for $K=1$ is outperformed when using more images, we can conclude that pairwise registration is not optimal when more images are available under moderate to high SNR levels (see figures~\ref{fig:comparMLEK} and~\ref{fig:comparMLEBoundsNoiseK}).

Similarly, as predicted by the EZZB for white signals, we observe a transition zone where performance degrades dramatically to finally converge to a flat zone. The flat region corresponds to SNR levels that are too low to enable alignment at all. For flat spectrum images and pairwise alignment ($K=1$), the EZZB accurately predicts the SNR threshold that defines the beginning of this transition region. However, for the multi-image case ($\K=10$) the \mleA{} algorithm performs worse than the prediction given by the EZZB (see Figure~\ref{fig:comparMLEBoundsNoiseK}). One reason for this might be that the initialization of the non-convex optimization in \mleA{} is performed using the pairwise registration to one of the input images, which is certainly not optimal. Nevertheless, as predicted by EZZB, the breaking point of the \mleA{} is pushed back several dBs when using more images in the registration (see figures~\ref{fig:comparMLEK} and~\ref{fig:comparMLEBoundsNoiseK}).

\medskip

\section{Conclusions}
\label{sec:discussion}

In this work, we analyzed the fundamental performance limits in image registration when multiple shifted and noisy observations are available. 
We derived and analyzed Cram\'er-Rao and extended Ziv-Zakai  bounds under different statistical models for both the underlying image and the shift vectors.

The first clear finding is that there is a \textit{per-region} behavior depending on the difficulty level of the problem, given by the SNR conditions (see for example figures~\ref{fig:comparBoundsK1} and~\ref{fig:comparMLEBoundsNoiseK}). At very high SNR, the performance is linear with both, the SNR and the image size, and it is independent of the prior information on the shifts and the number of available images. 
Indeed, all computed bounds agree, and the MLE achieves the bounds. Hence, doing pairwise alignment using the MLE gives the optimal performance. 

Assuming a stochastic image model, in high to moderate SNR scenarios, the performance is super-linear with the SNR and linear with the image size. 
Increasing the number of images widens the region where performance is linear with the SNR (very high SNR), so it improves registration. 
This is true for both considered stochastic image models: flat power spectral density  or with quadratic decay. 
Also, this agrees with the empirical MLE performance in this SNR range.

According to the computed extended Ziv-Zakai bound, there exists a critical SNR below which performance degrades dramatically with SNR. 
Having access to more images or increasing the image size help to push the SNR levels at which this transition zone starts. 
In very low SNR, the performance saturates to a value essentially given by the prior variance on the shifts. 
In this SNR region no alignment is possible.

In general, having access to more images improves the performance up to a certain limit. 
The exception is within the very high SNR region, where pairwise alignment is optimal.
Increasing the image size always improves performance, linearly reducing the performance bounds
and pushing the critical thresholds delimiting the transition and saturation zones. 
The studied shifts priors only had an impact at low SNR levels. 

As future work, we would like to analyze the impact of having more complex shift priors, for instance modeling correlation between the acquired frames (e.g., modeled by a random walk).
In addition, targeting a particular class of images, could help to develop better image priors. 
This will have an impact on the moderate to low SNR levels, since the performance in high SNR is found to be independent of the image prior.
Indeed, \mleA{}  has proven to be optimal when registering white noise signals (for the considered image size), but suboptimal for natural images in moderate SNR conditions.
Prior information could help to close the gap between the fundamental limit and the MLE performance.

\section*{Acknowledgments}
Work partially supported by the Department of Defense and NSF. 
The authors would like to thank Jean-Michel Morel for fruitful comments and discussions.
This work is in honor of Prof. Moshe Zakai, he will always be remembered as one of the greatest.

%Appendices
\appendices

\section{Cram\'er-Rao bound: deterministic image model}
\label{Ap:detCRB}
Let $\bz$ be $(\K+1)$ independent samples following~\eqref{eq:model}, and assuming $\bu$ is an unknown deterministic image, the log-likelihood function of $\bz$ with $\btheta=[\bu^T, \btau^T]^T$  is given by
\begin{equation}
\ell (\bz;\btheta) = - \frac{1}{2\sigma^2}\sum_{i=0}^{\K} ||\z_i(\bx) - \u(\bx - \btau_i)||^2,
\label{mleFuncAp}
\end{equation}
where we discarded the terms independent of $\btheta$. To compute the CRB we first compute the FIM
\begin{equation}
\mathbf{J}_D =  -\E_{\bz|\btheta} \left [ \frac{\partial^2 \ell(\bz ; \btheta )}{\partial \btheta^2} \right ] = 
      \begin{bmatrix}
	\mathbf{J}_{\bu\bu} & \mathbf{J}^T_{\bu\btau} \\      
        \mathbf{J}_{\bu\btau}  & \mathbf{J}_{\btau \btau} 
      \end{bmatrix}.
      \label{eq:fimOrigAp}
\end{equation}
Hence,
\begin{align}
&\mathbf{J}_{\bu\bu} =  -\E_{\bz|\btheta} \left [ \frac{\partial^2 \ell(\bz ; \btheta )}{\partial \bu^2} \right ]  = \frac{1}{\sigma^2}(\K+1) \I_{N_p}, \label{eq:JdApp1}\\
&\mathbf{J}_{\btau\btau} =  -\E_{\bz|\btheta} \left [ \frac{\partial^2 \ell(\bz ; \btheta )}{\partial \btau^2} \right ]  = \frac{1}{\sigma^2} \I_{\K} \otimes \mathbf{Q}, \label{eq:JdApp2} \\
&\mathbf{J}_{\bu\btau} =  -\E_{\bz|\btheta} \left [ \frac{\partial^2 \ell(\bz ; \btheta )}{\partial \bu \partial \btau} \right ]  = \frac{1}{\sigma^2}\oV \otimes [\bu^T_x, \bu^T_y]^T, \label{eq:JdApp3}
\end{align}
where $\I_{\Np}$ is the identity matrix of size $\Np \times \Np$ (idem for $\I_K$), $\oV$ is a vector of ones of size $\K$, $\otimes$ is the Kronecker product,  $\bu_x$, $\bu_y$ are the derivatives of the latent image $\bu$ in the horizontal and vertical directions respectively, and
\begin{equation}
\mathbf{Q} = 
      \begin{bmatrix}
	\bu_x^T\bu_x & \bu_x^T\bu_y \\      
        \bu_x^T\bu_y  & \bu_y^T\bu_y  
      \end{bmatrix}.
\end{equation}

Using the block matrix inversion principle~\cite{graybill2001}, the inverse of $\mathbf{J}_D$ can be expressed as
\begin{equation}
\mathbf{J}_D^{-1} = 
      \begin{bmatrix}
	\mathbf{S}_{\bu}^{-1} & \mathbf{J}^{-1}_{\bu\bu}\mathbf{J}_{\bu\btau}\mathbf{S}_{\btau}^{-1} \\      
        \mathbf{S}^{-1}_{\btau}\mathbf{J}_{\bu\btau}^T\mathbf{J}_{\bu\bu}^{-1}  & \mathbf{S}_{\btau}^{-1} 
      \end{bmatrix},
\label{eq:JdInvApp}      
\end{equation}
where
\begin{equation}
\bS_{\bu} = \bJ_{\bu\bu}  - \mathbf{J}_{\bu\btau} \mathbf{J}_{\btau \btau}^{-1} \mathbf{J}^T_{\bu\btau}, \quad \bS_{\btau} = \mathbf{J}_{\btau\btau}  - \mathbf{J}_{\bu\btau}^T \mathbf{J}_{\bu \bu}^{-1} \mathbf{J}_{\bu\btau}.
\label{eq:StauApp}
\end{equation}
From~\eqref{eq:JdApp1}--\eqref{eq:JdApp3} and~\eqref{eq:StauApp} we have,
\begin{equation}
\mathbf{S}^{-1}_{\btau} = \sigma^2 \left( \I_{\K} + \oV \oV^T \right) \otimes \mathbf{Q}^{-1}.                                     
\label{eq:StauInvApp}
\end{equation}
The Cram\'er-Rao bound in~\eqref{trBound} follows from \eqref{eq:StauInvApp}.

\section{Bayesian Cram\'er-Rao bound with shifts prior}
\label{app:BCRB}
Let $\bz$ be  $(\K+1)$ independent samples following model~\eqref{eq:model}, and assuming that $\bu$ is an unknown deterministic image and $\btau$ is a random variable following a generalized Gaussian prior $p (\btau)$ given by~\eqref{eq:GenGaus}, the joint log-likelihood $\ell (\bz, \btheta)$, with $\btheta = [\bu^T,\btau^T]^T$  is given by
\begin{equation}
\ell (\bz, \btau ; \bu) = \ln p(\bz | \btau;\bu) + \ln p (\btau).
\end{equation}
Hence, from~\eqref{eq:BFIM} the Bayesian FIM becomes 
\begin{equation}
\bJ_B \!=\! -\E_{\bz,\btau|\bu} \left[\frac{\partial^2 \ln p(\bz|\btau;\bu)}{\partial \btheta^2}\right] - \E_\btau \left[\frac{\partial^2 \ln p(\btau)}{\partial \btheta^2}\right].
\label{eq:Jb}
\end{equation}
Next, from~\eqref{eq:fimOrigAp},
\begin{align}
\!\!\!\E_{\bz,\btau|\bu} \left[\frac{\partial^2 \!\ln p(\bz|\btau;\bu)}{\partial \btheta^2}\right] \!&= \E_{\btau} \left[ \E_{\bz|\bu,\btau} \! \left[\frac{\partial^2 \!\ln p(\bz|\btau;\bu)}{\partial \btheta^2}\right] \right] \nonumber\\
&= -\E_{\btau} [ \bJ_D ] =  -\bJ_D.
\label{eq:JdPriorApp}															   		
\end{align}
%
%\begin{align}
%-\E_{\bz,\btau|\bu} \left[\frac{\partial^2 \ln p(\bz|\btau;\bu)}{\partial \btheta^2}\right] &= \E_{\btau} [ \bJ_D ] \\
%															   &= \bJ_D.
%\label{eq:JdPriorApp}															   		
%\end{align}
%
For the generalized prior~\eqref{eq:GenGaus}, 
%Also, for a Generalized Gaussian prior~\eqref{eq:GenGaus}, 
it can be shown that~\cite{nguyen94}
\begin{align}
-\E_{\btau} \left [ \frac{\partial^2 \ln p(\btau)}{\partial \btau^2} \right ] &= \frac{1}{\lm} \mathbf{I}_{2\K},
%														  &\mydef \frac{1}{\lm} \mathbf{I}_{2\K}.
%-\E_{\btau} \left [ \frac{\partial^2 \ln p(\btau)}{\partial \btau^2} \right ] &\phantom{:}= \frac{c^2\Gamma(3/c)\Gamma(2-1/c)}{\delta^2 \Gamma^2(1/c)} \mathbf{I}_{2\K}\\
%														  &\mydef \frac{1}{\lm} \mathbf{I}_{2\K}.
\label{eggApp}
\end{align}
where $\lambda^2 \mydef \frac{\delta^2 \Gamma^2(1/c)}{c^2\Gamma(3/c)\Gamma(2-1/c)}$.
Hence, from~\eqref{eq:Jb}--\eqref{eggApp},
\begin{equation}
\bJ_B = \bJ_D + \mathbf{J}_p, \text{ with } \quad 
\mathbf{J}_p = 
      \begin{bmatrix}
	0 & 0 \\      
        0  & \frac{1}{\lm} \mathbf{I}_{2\K}
      \end{bmatrix}.
\end{equation}
Using the block matrix inversion principle~\cite{matrixCookbook2012}, the inverse of $\mathbf{J}_B$ can be expressed as
\begin{equation}
\mathbf{J}_B^{-1} = 
      \begin{bmatrix}
	\bar{\mathbf{S}}_{\bu}^{-1} & \mathbf{J}^{-1}_{\bu\bu}\mathbf{J}_{\bu\btau}\mathbf{S}_{\btau}^{-1} \\      
        \mathbf{S}^{-1}_{\btau}\mathbf{J}_{\bu\btau}^T\mathbf{J}_{\bu\bu}^{-1}  & \bar{\mathbf{S}}_{\btau}^{-1} 
      \end{bmatrix},
\label{eq:JdInv2App}      
\end{equation}
with
\begin{align}
&\bar{\bS}_{\bu} =  \bJ_{\bu\bu}  - \mathbf{J}_{\bu\btau} \left(\mathbf{J}_{\btau \btau} + \tfrac{1}{\lm} \mathbf{I}_{2\K} \right)^{-1} \mathbf{J}^T_{\bu\btau} \\
&\bar{\bS}_{\btau} = \mathbf{J}_{\btau\btau}  + \tfrac{1}{\lm} \mathbf{I}_{2\K} - \mathbf{J}_{\bu\btau}^T \mathbf{J}_{\bu \bu}^{-1} \mathbf{J}_{\bu\btau}. 
\label{eq:StaubApp}
\end{align}
where $\mathbf{J}_{\bu \bu}$, $\mathbf{J}_{\btau\btau}$ and  $\mathbf{J}_{\bu\btau}$ are given by~\eqref{eq:JdApp1}--\eqref{eq:JdApp3}. 
Hence,
%Hence, from~\eqref{eq:JdApp1}--\eqref{eq:JdApp3} and~\eqref{eq:StaubApp},
\begin{equation}
\bar{\bS}_{\btau} = \tfrac{1}{\ss}\left (  \mathbf{I} - (K+1) \oV \oV^T \right ) \otimes \mathbf{Q} + \tfrac{1}{\lm} \mathbf{I},
\label{eq:Stau2bApp}
\end{equation}
and
\begin{align}
\bar{\mathbf{S}}_{\btau}^{-1} &= \mathbf{I} \otimes \left ( \tfrac{1}{\ss} \mathbf{Q} + \tfrac{1}{\lm} \I \right )^{-1}  \nonumber\\
                        &+ \oV \oV^T\! \otimes \! \lambda^2 \left ( (\K\! +\! 2) \I + \tfrac{\lm}{\ss} \mathbf{Q} + (\K\! +\!1) \tfrac{\ss}{\lm} \mathbf{Q}^{-1} \right )^{-1}.    
\label{eq:StauInvBApp}                                            
\end{align}
The Bayesian Cram\'er Rao bound in~\eqref{trBoundBayes} follows from \eqref{eq:StauInvBApp}.

\section{Cram\'er-Rao bound: Stochastic image model}
\label{app:crlbRand}
Let $\tilde{\bz}$ be given by \eqref{eq:zFourier}.
%
%Let us consider $\hat{\bz}$ given by the stochastic model in \eqref{}, that is 
%%
%\hat{\bz} follows a complex Gaussian distribution having zero mean and covariance matrix $\bSigma$ given in $in \eqref{eq:Kdef}$$
%%
This random variable follows a complex Gaussian distribution with zero mean and covariance matrix $\bSigma$ given by~\eqref{eq:Kdef}.
The FIM corresponding to the complex Gaussian process $\hat{\bz}$ is given by \cite[Ap. 15C]{kay1993fundamentals}
\begin{equation}
\{\bJ_S\}_{i_h,j_q} =\sum_{l=1}^M \tr \left ( \bSigma_\btau^{-1}(\bomega_l) \frac{\partial \bSigma_\btau(\bomega_l)}{\partial \t_{i_h}} \bSigma_\btau^{-1}(\bomega_l) \frac{\partial \bSigma_\btau(\bomega_l)}{\partial \t_{j_q}} \right ),
\label{eq:JstochApp}
\end{equation}
%\begin{align}
%\{\bJ_S\}_{i_h,j_q} &= \tr \left ( \bSigma^{-1} \frac{\partial \bSigma}{\partial \t_{i_h}} \bSigma^{-1} \frac{\partial \bSigma}{\partial \t_{j_q}} \right )\\
%              &=  \sum_{l=1}^M \tr \left ( \bSigma_\btau^{-1}(\bomega_l) \frac{\partial \bSigma_\btau(\bomega_l)}{\partial \t_{i_h}} \bSigma_\btau^{-1}(\bomega_l) \frac{\partial \bSigma_\btau(\bomega_l)}{\partial \t_{j_q}} \right ),
%\label{eq:JstochApp}
%\end{align}
where $i,j=1,\dotsc,\K$, and $h,q \in \{x,y\}$ index the two components of each 2D shift vector $\btau_i$. The spatial frequency $\bomega_l$, $l(l_x,l_y)=1,\dotsc,M$ with $M=m_c + \frac{m_r}{2} + 2$ indexes the 2D frequencies $\bomega_{l} =[\omega_{l_x},\omega_{l_y}]^T$ with $\omega_{l_x} = \frac{2\pi l_x}{m_c}, l_x = -\frac{m_c}{2},\dotsc,\frac{m_c}{2}$ and $\omega_{l_y} = \frac{2\pi l_y}{m_r}, l_y = 0,\dotsc,\frac{m_r}{2}$. To simplify notation, we avoid in the following the subindex $l$ on $\bomega$.  The matrix $\bSigma_\btau(\bomega)$ can be decomposed as
\begin{equation}
\bSigma_\btau(\bomega) = S(\bomega) \bP_\btau(\bomega) \bP_\btau(\bomega)^H + N(\bomega) \I_{K+1},
\label{eq:sigmaApp}
\end{equation}
with
\begin{equation}
\bP_\btau(\bomega) = [1, e^{i \bomega \cdot \btau_1}, e^{i \bomega \cdot \btau_2}, \ldots, e^{i \bomega \cdot \btau_K}]^T.
\end{equation}
%and $\I$ is the identity matrix of size $(K+1)\times(K+1)$.
%
Using the Sherman-Morrison formula~\cite{matrixCookbook2012},
\begin{equation}
\bSigma_\btau^{-1}(\bomega) = N^{-1}(\bomega) \Big( \bI_{K+1} + \alpha(\bomega) \bP_\btau(\bomega) \bP_\btau(\bomega)^H \Big),
\label{eq:sigmaInvApp}
\end{equation}
where 
\begin{equation}
\alpha(\bomega)= -\frac{S(\bomega)}{N(\bomega) + (K+1)S(\bomega)}.
\end{equation}
To simplify notation we avoid in the following the dependence on $\bomega$.
Hence we have
\begin{equation}
\bSigma_\btau^{-1} \frac{\partial \bSigma_\btau}{\partial \t_{i_h}} = N^{-1} \left(  \frac{\partial \bSigma_\btau}{\partial \t_{i_h}} + \alpha \bP_\btau \bP_\btau^H  \frac{\partial \bSigma_\btau}{\partial \t_{i_h}}\right),
\end{equation}
and 
\begin{align}
&\tr \left ( \bSigma_\btau^{-1} \frac{\partial \bSigma_\btau}{\partial \t_{i_h}} \bSigma_\btau^{-1} \frac{\partial \bSigma_\btau}{\partial \t_{j_q}} \right ) = \nonumber\\
&\quad  N^{-1} \left [ \tr \left( \frac{\partial \bSigma_\btau}{\partial \t_{i_h}}  \frac{\partial \bSigma_\btau}{\partial \t_{j_q}} \right)  +  2\alpha \tr \left( \frac{\partial \bSigma_\btau}{\partial \t_{i_h}} \bP_\btau \bP_\btau^H \frac{\partial \bSigma_\btau}{\partial \t_{j_q}} \right) \right. \nonumber\\
&\phantom{=} + \left. \alpha^2 \tr \left(  \bP_\btau \bP_\btau^H \frac{\partial \bSigma_\btau}{\partial \t_{i_h}}  \bP_\btau \bP_\btau^H \frac{\partial \bSigma_\btau}{\partial \t_{j_q}} \right) \right ].
\label{eq:auxTrace}
\end{align}
Substituting \eqref{eq:auxTrace} in \eqref{eq:JstochApp} and computing the derivatives, 
\begin{equation}
%\bJ = \rho (\oV \oV^T - (\K + 1) \I),
\bJ_S = \left[ (\K + 1) \I_{\K} - \oV \oV^T \right] \otimes \mathbf{B}, \text{ with }
\mathbf{B} = 
      \begin{bmatrix}
	\rho_{x,x} &\!\! -\rho_{x,y} \\      
        -\rho_{x,y}  &\!\! \rho_{y,y}
      \end{bmatrix},
\end{equation}
\begin{equation}
%\rho_{h,q} = 2\sum_{l=1}^M \frac{S^2(\bomega_l)}{N^2(\bomega_l)} \omega_{l_h} \omega_{l_q} [1 + \alpha(\bomega_l)(\K + 1)].
\rho_{h,q} = \sum_{l=1}^M  \frac{2S^2(\bomega_l)\, \omega_{l_h} \omega_{l_q}}{N^2(\bomega_l) + (K+1)S(\bomega_l)N(\bomega_l)}.
\label{eq:gammaHQApp}
%
%\rho = 2\sum_{l=1}^M \frac{S^2(\bomega_l)}{N^2(\bomega_l)} \bomega_l^2 [1 + \alpha(\bomega_l)(\K + 1)] ,
\end{equation}
Hence, we have
\begin{equation}
\bJ_S^{-1} = \tfrac{1}{(\K+1)} \left( \I_K + \oV \oV^T \right) \otimes \mathbf{B}^{-1}.
\end{equation}
Then, the error on the shifts estimates is bounded by 
\begin{equation}
\mse \geq \frac{1}{2\K}\tr(\bJ_S^{-1}) = \frac{1}{(\K+1)} \left( \frac{\rho_{x,x}^2 + \rho_{y,y}^2 }{\rho_{x,x} \rho_{y,y} - \rho_{x,y}^2 } \right).
\label{wq:bound2Ap}
\end{equation}
Notice that if $S(\bomega)$ and $N(\bomega)$ are rotational symmetric (rotation invariant), we have
%\begin{align}
%\rho_{x,y} &= 2\sum_{l=1}^M \frac{S^2(\bomega_l)}{N^2(\bomega_l)} \omega_{l_x} \omega_{l_y} [1 + \alpha(\bomega_l)(\K + 1)] \\
%		      &= 2\sum_{l_x=\frac{-m_c}{2}}^{\frac{m_c}{2}}  \omega_{l_x} \sum_{l_y=0}^{\frac{m_r}{2}} \omega_{l_y} \frac{S^2(\omega_{l_x},\omega_{l_y})}{N^2(\omega_{l_x},\omega_{l_y})} [ 1 \\ 
%		      &\phantom{=} \qquad \qquad \qquad \qquad + \alpha(\omega_{l_x},\omega_{l_y})(\K + 1)] \\
%		      &= 0,
%\end{align}
%and $\rho_{x,x} = \rho_{y,y}$.
$\rho_{x,y}=0$ and $\rho_{x,x} = \rho_{y,y}$.
In that case, the CRB bound on the MSE~\eqref{wq:bound2Ap} becomes
\begin{equation}
%\crbs \mydef  \frac{1}{(\K+1)} \left [ \frac{\rho_{x,x}^2 + \rho_{y,y}^2 }{\rho_{x,x} \rho_{y,y} } \right ].
\crbs \mydef  \frac{2}{(\K+1) \rho_{xx}} = \frac{2}{(\K+1) \rho_{yy}}. %\left [ \frac{\rho_{x,x}^2 + \rho_{y,y}^2 }{\rho_{x,x} \rho_{y,y} } \right ].
\label{wq:boundAp}
\end{equation}

\medskip

\noindent \textbf{High SNR performance.}
When the signal-to-noise ratio is very high, i.e., $S(\bomega)/N(\bomega) \gg 1$, we have that
\begin{equation}
\rho_{x,x} \xrightarrow{S/N \rightarrow \infty} \rho^\textsc{hsnr}_{x,x} = \sum_{l=1}^M  \frac{2S(\bomega_l)\, \omega^2_{x}}{(K+1)N(\bomega)}.
\label{eq:GammaHSNR}
\end{equation}
Let us assume the noise is white with $N(\bomega) = \sigma^2$. 
Since,
\begin{align}
\sum_{l=1}^M  S(\bomega_l)\, \omega^2_{l_x} &\approx  \frac{N_p}{(2 \pi)^2} \int_{0}^{\pi} \int_{-\pi}^{\pi} S(\bomega) \omega^2_x \diff \omega_x \diff \omega_y,\\
&=  \frac{1}{2}\frac{N_p}{(2 \pi)^2}  \int_{-\pi}^{\pi}\int_{-\pi}^{\pi} S(\bomega) \omega^2_x  \diff \omega_x \diff \omega_y.
\end{align}
Then, the CRBS  for a rotation invariant process (Eq. \eqref{wq:boundAp}) under high SNR simplifies to
\begin{equation}
\crbs^\textsc{hsnr} \mydef  \frac{2\sigma^2 (2 \pi)^2}{ N_p \int S(\bomega) \omega^2_x \diff \bomega} . %\left [ \frac{\rho_{x,x}^2 + \rho_{y,y}^2 }{\rho_{x,x} \rho_{y,y} } \right ].
\label{eq:crbsHSNR}
\end{equation}
This bound is independent of the number of images $K$. 
As we show at follows, it agrees with the deterministic CRB.

Let $\bu$ be a deterministic image with $N_p\gg1$ pixels, that we assume rotation invariant for simplicity. 
We can approximate the power spectral density by its empirical power spectrum $S_d(\bomega)$. Then,
\begin{equation}
\frac{1}{N_p} \bu_x^T \bu_x = \frac{1}{(2\pi)^2 }\int S_d(\bomega) \omega^2_x \diff \bomega. 
\end{equation}
For a rotation invariant image $\bu$, we have that 
$$\bu_x^T\bu_y = \frac{1}{(2\pi)^2}\int \tilde{\bu}(\omega_x,\omega_y) \omega_x \omega_y \diff \omega_x \diff \omega_x  = 0.$$

Next, we have that the deterministic CRB in~\eqref{trBound}, for a rotation invariant signal, can be rewritten as
\begin{align}
\crbd = \frac{2\sigma^2}{ \bu_x^T \bu_x} =  \frac{2\sigma^2 (2\pi)^2}{N_p \int S_d(\bomega) \omega^2_x \diff \bomega}.
\end{align}
That is, the stochastic CRB bound in high SNR agrees with the deterministic CRB. 

\medskip

\noindent \textbf{Flat Spectrum signals.}
Let us consider the particular case of flat spectrum signals, that is,
\begin{equation}
S(\bomega ) = 
\begin{cases}
S_w & \text{if }  \max(|\omega_x|,|\omega_y|)\le  W/2,\\
0 & \text{otherwise.}
\end{cases}
\end{equation}
We will also assume that the additive noise spectrum is flat in the same frequency band $[-\frac{W}{2}, \frac{W}{2}]^2$ and zero otherwise. 
In this case, if we assume $M\gg1$, we can consider the sum in~\eqref{eq:gammaHQApp} for $\rho_{xx}$, as a Riemann approximation, that is,
\begin{align}
%&\hspace{-3em}\sum_{l=1}^M \frac{2S(\bomega_l)^2 \omega_{l_x}^2}{N^2(\bomega_l) \!+\! (\K \!+\! 1)S(\bomega_l) N(\bomega_l)}  \nonumber\\
\rho_{xx}&\approx \frac{2N_p}{(2\pi)^2}\int_{0}^{\frac{W}{2}}\! \!\!\int_{0}^{\frac{W}{2}} \!\! \frac{ 2S^2 \omega^2_{l_x}}{N^2 \!+\! (\K\! +\! 1) S N}  \diff \bomega \\
%											   &= \frac{N}{2\pi} \frac{\snr^2}{1 + (\K + 1)\snr}  \int_0^{W/2}   \omega ^2 \diff \omega \\
&\quad= \frac{S^2 W^4 N_p}{3\pi^2 2^4(N^2 + (\K + 1)S N)}.
\label{eq:SumWhiteApp}
\end{align}
Thus, rewriting~\eqref{eq:SumWhiteApp}  in terms of the SNR as defined in \eqref{eq:SNRW}, we obtain that the CRB (Eq.~\eqref{wq:boundAp}) for  white images
is: 
%%1D
%\begin{equation}
%\crbsw \mydef \frac { \pi \left( W^2 + 12 (K+1)\snrg_w \right)}{3(K+1)\Np W\snrg_w^2}.
%\end{equation}
\begin{equation}
\crbsw \mydef \frac{8 \pi^2 \left( W^2 + 6 \snrg_w(K+1) \right)}{\Np(K+1)\snrg_w^2 3 W^2}.
\end{equation}

\noindent \textbf{Natural images.} 
One classical assumption when modeling natural images is that the power spectrum falls
quadratically with the Fourier frequency.  Let us assume that the considered underlying image follows this law, that is,
\begin{equation}
S(\bomega ) = 
\begin{cases}
 S_n \|\bomega \|^{-2} & \text{if} \quad  \max(|\omega_x|,|\omega_y|)\le  W/2,\\
0 & \text{otherwise.}
\end{cases}
\label{eq:whiteSignalAp}
\end{equation}
Similarly as for the white signals, let us assume that the additive noise spectrum is flat in the same frequency band $[-\frac{W}{2}, \frac{W}{2}]^2$ taking value $N$ and zero otherwise.
We can approximate the sum in $\rho_{xx}$ (Eq.~\eqref{eq:gammaHQApp}) by,
%\begin{align}
%& \frac{(2\pi)^2}{N_p}\sum_{l=1}^M \frac{2S^2(\bomega_l)  \omega^2_{l_x}}{N(\bomega_l)^2 \!+\! (\K \!+\! 1)S(\bomega_l) N(\bomega_l)} \nonumber\\
%&\quad \approx 2\int_{0}^{\frac{W}{2}} \!\!\!\!\!\int_0^{\frac{W}{2}} \!\!\frac{2S_n^2 \omega_x^2}{N^2(\omega^2_x \!+\! \omega^2_y)^2 \!+\! (\K \!+\! 1)S_n  N (\omega^2_x \!+\! \omega^2_y)} \diff \bomega.
%\label{eq:intNaturalXApp}
%\end{align}
%
\begin{align}
%&\sum_{l=1}^M \frac{2S^2(\bomega_l)  \omega^2_{l_x}}{N(\bomega_l)^2 \!+\! (\K \!+\! 1)S(\bomega_l) N(\bomega_l)} \nonumber\\
\rho_{xx} & \approx \frac{2N_p}{(2\pi)^2}\!\! \int_{0}^{\frac{W}{2}} \!\!\!\!\!\!\int_0^{\frac{W}{2}} \!\!\frac{2(\tfrac{S_n}{N})^2 \omega_x^2}{(\omega^2_x \!+\! \omega^2_y)^2 \!+\! \tfrac{S_n}{N}(\K \!+\! 1) (\omega^2_x \!+\! \omega^2_y)} \diff \bomega.
\label{eq:intNaturalXApp}
\end{align}
Due to symmetry,
\begin{align}
& \int_{0}^{\frac{W}{2}} \!\!\!\!\!\int_0^{\frac{W}{2}} \!\!\frac{2S_n^2 \omega_x^2}{N^2(\omega^2_x \!+\! \omega^2_y)^2 \!+\! (\K \!+\! 1)S_n  N (\omega^2_x \!+\! \omega^2_y)} \diff \bomega\\
&\,= \int_{0}^{\frac{W}{2}} \!\!\!\!\!\int_0^{\frac{W}{2}} \!\!\frac{S_n^2(\omega_x^2+\omega_y^2)}{N^2(\omega^2_x \!+\! \omega^2_y)^2 \!+\! (\K \!+\! 1)S_n  N (\omega^2_x \!+\! \omega^2_y)} \diff \bomega \label{eq:approxQuartCircle}\\
&\,\approx  \frac{\pi}{2} \int_{0}^{\frac{W}{\sqrt{\pi}}} \frac{S_n^2 r}{N^2r^2 + (K+1)S_nN} \diff r \\
&\, = \tfrac{\pi}{2} S^2_n \acoth \left( 1 + \tfrac{2\pi(K+1)S_n}{W^2 N } \right).
\label{eq:intNaturalXYApp}
\end{align}
The approximation in~\eqref{eq:approxQuartCircle} is done by  changing the area of integration from $[0,\frac{W}{2}]^2$ to  the quarter of circle $[0,\tfrac{\pi}{2}]$ of radius $\frac{W}{\sqrt{\pi}}$. This is the maximum overlapping circular region that covers the same area as the original one. 
Thus, under the considered natural image model, the CRB in~\eqref{wq:boundAp} can be approximated by
%\begin{equation}
%\crbsn \mydef 8 \pi \left(N_p (K+1)  \snr_n^2  \acoth \left( 1 \!+\! \frac{8(K\!+\!1)\snr_n}{W^2} \right) \right)^{-1},
%\end{equation}
%
\begin{equation}
\crbsn \mydef \frac{8 \pi}{N_p (K+1)  \snr_n^2  \acoth \left( 1 \!+\! \frac{2\pi(K\!+\!1)\snr_n}{W^2} \right)},
\end{equation}

where $\snr_n$ is defined in \eqref{eq:SNRnat}.

\section{Extended Ziv-Zakai Bound: Probability of error}
\label{sec:probErrorEZZB}
The computation of the Extended Ziv-Zakai bound requires computing the probability of error  $P^\text{el}_\text{min} (\bphi, \bphi + \bdelta)$, for the equally likely hypothesis case.
This probability can be tightly approximated~\cite[Eq. (2.243)]{vanTrees2013detection} by
{\small
\begin{align}
&P^\text{el}_\text{min} (\bphi, \bphi + \bdelta) \approx \\ \nonumber
&\,\, \tfrac 12 \exp \left\{ \mu(s_m) + \tfrac{s^2_m}{2} \mu''(s_m) \right\} \Phi \left(s_m\sqrt{ \mu''(s_m)}\right)+\\
&\quad \tfrac 12 \exp \left\{ \mu(s_m) + \tfrac{(s_m\!-\!1)^2}{2} \mu''(s_m) \right\} \Phi \left((1\!-\!s_m)\sqrt{ \mu''(s_m)}\right) \nonumber
\end{align}
}
where
\begin{equation}
\mu(s) = \log \int \left[ p(\tilde{\bz} \cond \bphi) \right]^s \left[p(\tilde{\bz} \cond \bphi + \bdelta) \right]^{1-s} \diff \tilde{\bz},
\end{equation}
$s_m$ verifies $\mu'(s_m) = 0$ and 
$\Phi(t) = \tfrac{1}{\sqrt{2 \pi}}\int_t^\infty e^{-\frac{t^2}{2}} \diff t$.

Assuming that the Fourier coefficients at different frequencies are statistically uncorrelated this becomes
\begin{multline}
\mu(s) = - \sum_{l=1}^M \Big\{ s \log \big \lvert \bSigma_{\btau + \bdelta} (\bomega_l) \big \rvert  + (1- s) \log \big \lvert \bSigma_{\btau} (\bomega_l) \big\rvert \\
 +   \log \left( \big \lvert s \bSigma^{-1}_{\btau + \bdelta}(\bomega_l) + (1-s) \bSigma^{-1}_{\btau}(\bomega_l)  \big \rvert \right)  \Big\},
\label{eq:muGaussAp}
\end{multline}
where $\bSigma_\btau(\bomega)$ is defined in~\eqref{eq:sigmaApp}.
By doing some algebra manipulations one can see that
{\small
\begin{equation}
\big \lvert \bSigma_{\btau}(\bomega)  \big \rvert  \!=\! \big \lvert \bSigma_{\btau + \bdelta}(\bomega)  \big \rvert \!=N(\bomega)^{K}  \left( N(\bomega) + (K\!+\!1) S(\bomega) \right).
\label{eq:detKAp}
\end{equation}
}
%\mdC{Using the Sherman-Morrison formula,
%\begin{equation}
%\bSigma^{-1}_{\btau}(\bomega) = N^{-1}(\bomega) \Big( I + \alpha(\bomega) \bP_\btau(\bomega) \bP_\btau(\bomega)^H \Big),
%\end{equation}
%where 
%\begin{equation}
%\alpha(\bomega)= -\frac{S(\bomega)}{N(\bomega) + (K+1)S(\bomega)}.
%\end{equation}}
Next, from \eqref{eq:sigmaInvApp} we can rewrite the determinant in the second term of \eqref{eq:muGaussAp} as,
\begin{align}
&\lvert s \bSigma^{-1}_{\btau + \bdelta}(\bomega_l) + (1-s) \bSigma^{-1}_{\btau}(\bomega_l)  \big \rvert \nonumber\\
%&\quad=\Big \lvert  N(\bomega)^{-1}\Big(  I + s \alpha(\bomega) \bP_{\btau + \bdelta} (\bomega) \bP_{\btau + \bdelta} (\bomega)^H \nonumber\\
%&\qquad\qquad\qquad\qquad+ (1-s) \alpha(\bomega)  \bP_{\btau} (\bomega) \bP_{\btau } (\bomega)^H \Big) \Big \rvert \\ 
  &\quad= N(\bomega)^{-(K+1)} \Big(1 + s \alpha(\bomega) (K+1)\Big) \nonumber \\
  &\quad \quad \cdot\Big( 1 + (1-s) \alpha(\bomega) \big( (K+1) - \beta(\bomega) T(\bdelta,\bomega)\big)\Big),
\label{detKpKAp}
\end{align}
where 
{\small
\begin{equation}
\beta(\bomega) = \frac{s \alpha(\bomega)}{1\! +\! (K\!+\!1)s\alpha(\bomega)}, \,\, T(\bdelta,\bomega) = \Big\lvert1 + \sum_{j=1}^K e^{-i \bdelta_j \cdot \bomega}\Big\rvert^2.
\end{equation} 
}
%and 
%\begin{equation}
%T(\bdelta,\bomega) = \Big\lvert1 + \sum_{j=1}^K e^{-i \bdelta_j \cdot \bomega}\Big\rvert^2.
%\end{equation}
%
Thus substituting \eqref{eq:detKAp} and \eqref{detKpKAp} in \eqref{eq:muGaussAp} one obtains,
\begin{equation}
\mu(s) = -\sum_{l=1}^M \log \Big[ 1 + 4 s(1-s)  \gamma(\bdelta, \bomega_l) \Big],
\end{equation}
where 
\begin{equation}
\gamma(\bdelta, \bomega_l) = \frac{ S(\bomega_l)^2  \Big( (K+1)^2 - T(\bdelta,\bomega_l) \Big)}{4N(\bomega_l)^2 + (K+1)N(\bomega_l)S(\bomega_l)}.
\end{equation}
Thus,
\begin{equation}
\mu'(s) = \sum_{l=1}^M \frac{ 4\gamma(\bdelta, \bomega_l) (2s-1)} { 1+ 4s(1-s)\gamma(\bdelta, \bomega_l)}
\end{equation}
and the point such that $\mu'(s_m) = 0$ is $s_m=1/2$.
Then
{\small
\begin{equation}
\mu(\tfrac{1}{2}) =\! -\sum_{l=1}^M \log \left( 1\! +\! \gamma(\bdelta, \bomega_l) \right),\,\, \mu''(\tfrac{1}{2}) = \sum_{l=1}^M \frac{8\gamma(\bdelta, \bomega_l)}{1+\gamma(\bdelta, \bomega_l)}.
\end{equation}
}
%and
%\begin{equation}
%\mu''(\tfrac{1}{2}) = \sum_{l=1}^M \frac{8\gamma(\bdelta, \bomega_l)}{1+\gamma(\bdelta, \bomega_l)}.
%\end{equation}
%Since $\bomega_l = \frac{2\pi}{N_p}l$ and we are assuming $N_p\gg1$, then
%\begin{align}
% \mu(1/2) &\approx - \frac{N_p}{2\pi} \int_0^{W/2} \log \Big [ 1 + \gamma(\bdelta,\bomega) \Big] \diff \bomega\\
% \mu''(1/2) &\approx \frac{N_p}{2\pi} \int_0^{W/2} \frac{\gamma(\bdelta,\bomega)}{1+\gamma(\bdelta,\bomega)}  \diff \bomega.
%\end{align}

Finally, the probability of error can be approximated by
\begin{align}
%P^\text{el}_\text{min} (\bdelta) &= P^\text{el}_\text{min} (\btau, \btau + \bdelta)  \nonumber\\
P^\text{el}_\text{min} (\btau, \btau + \bdelta) \approx \tfrac 12 \exp \left\{ a(\bdelta) + b(\bdelta) \right\} \Phi \left(\sqrt{ 2 b(\bdelta)}\right),
 \label{eq:peAp}
\end{align}
where 
{\small
\begin{equation}
a(\bdelta) \!=\!  -\sum_{l=1}^M \log \left( 1 \!+\! \gamma(\bdelta, \bomega_l) \right), \,\,
b(\bdelta) \!=\!    \sum_{l=1}^M \frac{\gamma(\bdelta, \bomega_l)}{1\!+\!\gamma(\bdelta, \bomega_l)}, \label{eq:PeabOrigAp}
\end{equation} 
}
%
%\begin{equation}
%\gamma(\bdelta, \bomega) = \frac{ S(\bomega)^2  \Big( (K+1)^2 - T(\bdelta,\bomega) \Big)}{4\left( N^2(\bomega) + (K+1)N(\bomega) S(\bomega)\right)}
%\end{equation}
%and
%\begin{equation}
%T(\bdelta,\bomega) =\Big\lvert1 + \sum_{j=1}^K e^{-i \bdelta_j \cdot \bomega}\Big\rvert^2.
%\end{equation}

\section{Extended Ziv-Zakai bound for white signals}
\label{app:ezzbWS}
\setcounter{figure}{0} \renewcommand{\thefigure}{A.\arabic{figure}}
To simplify the analysis we consider 
one-dimensional random signals with a constant power spectral density $S(\omega)$ in 
$[-\frac{W}{2}, \frac{W}{2}]$ and zero otherwise, and zero-mean Gaussian noise with flat spectrum $N(\omega) = N$ in the same frequency band.
We closely follow the deduction in~\cite{weinstein1984fundamental}.
%To simplify the analysis we shall analyze the extended Ziv-Zakai lower bound for the particular case of 
%one-dimensional random signals having a constant compact support power spectral density $S(\omega)$ in 
%$[-\frac{W}{2}, \frac{W}{2}]$ and zero otherwise.
%We also assume that the Gaussian additive noise  also has a flat spectrum $N(\omega) = N$ in the same frequency band.
%We tightly follow the deduction in~\cite{weinstein1984fundamental}.

First, assuming $N_p\gg1$, let us approximate $a(\bdelta)$ and $b(\bdelta)$ in~\eqref{eq:PeabOrigAp} for the particular case of white signals, 
\begin{align}
a(\bdelta) &=  - \frac{N_p}{2\pi} \int_0^{W/2} \log \Big [ 1 + \gamma(\bdelta,\omega) \Big] \diff \omega,  \label{eq:PeaAp}\\
b(\bdelta) &=    \frac{N_p}{2\pi} \int_0^{W/2} \frac{\gamma(\bdelta, \omega)}{1+\gamma(\bdelta, \omega)} \diff \omega \label{eq:PebAp},
\end{align} 
and 
\begin{equation}
\gamma(\bdelta, \omega) = \frac{ S^2  \big( (K+1)^2 - T(\bdelta,\omega) \big)}{4\left( N^2 + (K+1)N S\right)}.
\label{eq:gamma}
\end{equation}
%This approximation is reasonable if $N_p \gg1$.

To evaluate the Extended Ziv-Zakai bound in estimating one single component of $\btau = [\tau_1,\tau_2,\ldots,\tau_M]$, we can choose without loss of generality $\ba = [1,0,\ldots,0]$ in \eqref{eq:ezzbSimpl}. 
Thus, evaluation of the EZZB requires computing
%\begin{equation}
%\epsilon_1^2 \ge \int_0^\infty \nuC \left\{ \max_{\bdelta:\delta_1 = h} A(\bdelta) P^\text{el}_\text{min} (\bdelta) \right\} h \diff h.
%\label{eq:ezzb1Ap}
%\end{equation}
\begin{equation}
\epsilon_1^2 \ge \int_0^\infty P_A(h) \diff h,
\label{eq:ezzb1Ap}
\end{equation}
where we have defined
\begin{equation}
P_A(h) = \nuC \left\{ \max_{\bdelta:\delta_1 = h} A(\bdelta) P^\text{el}_\text{min} (\bdelta) \right\}.
\end{equation}

Note that to solve \eqref{eq:ezzb1Ap} one needs to maximize
\begin{equation}
g(\bdelta) = A(\bdelta)P^\text{el}_\text{min}(\bdelta)
\label{eq:gApp}
\end{equation}
with respect to $[\delta_2,\ldots,\delta_K]$ for each value of $\delta_1=h$. A lossy (in general) lower bound can be obtained by setting the unspecified components of $\bdelta$ to zero.
Due to the symmetry of the problem it is clear that the maximum should be attained at $\bdelta$ such that $\delta_2=\delta_3=\ldots=\delta_K$. 
Thus, to simplify the exposition, let us do an abuse of notation and refer to $\bdelta$ as the couple $[\delta_1,\delta_2]$, and omit $\delta_3,\ldots,\delta_K$ assuming that they are all equal to $\delta_2$.

\begin{figure}
\centering
\includegraphics[width=.49\linewidth]{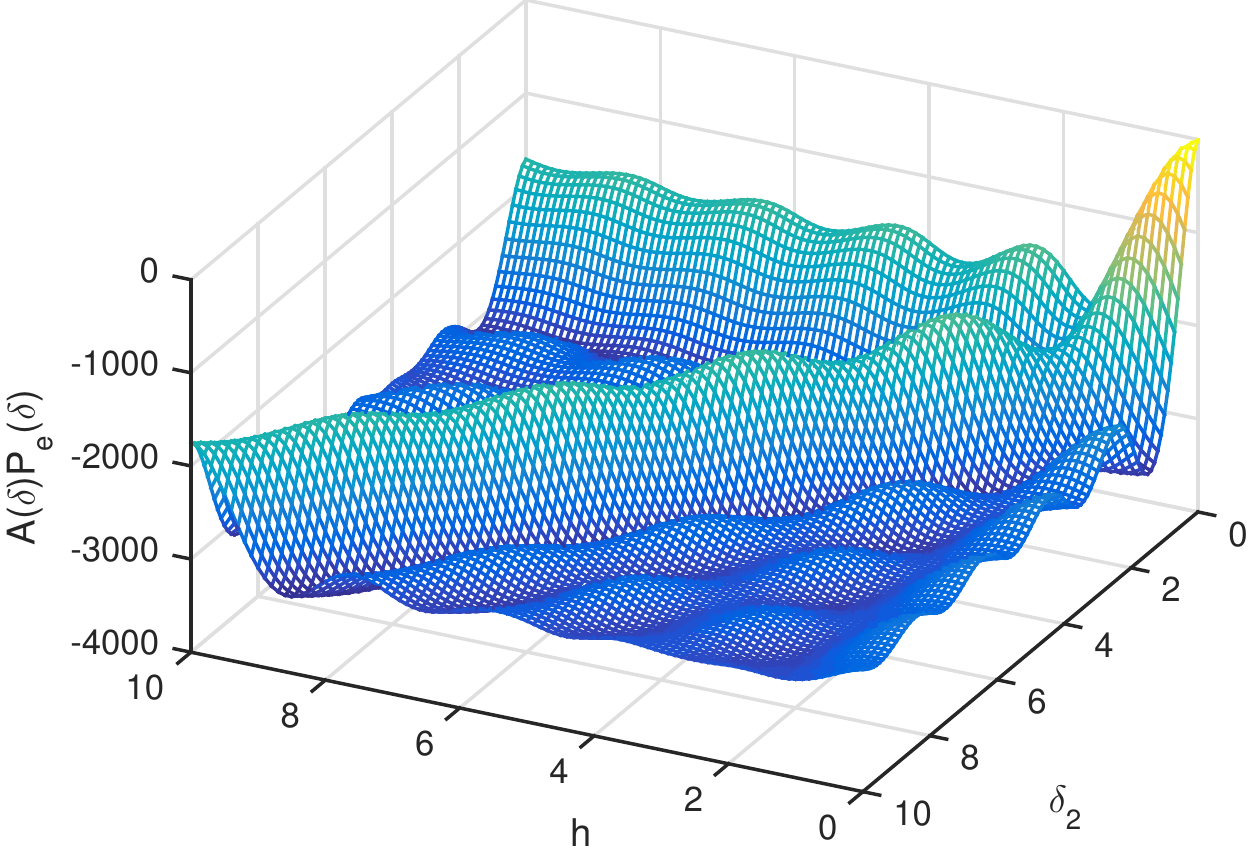}
\includegraphics[width=.49\linewidth]{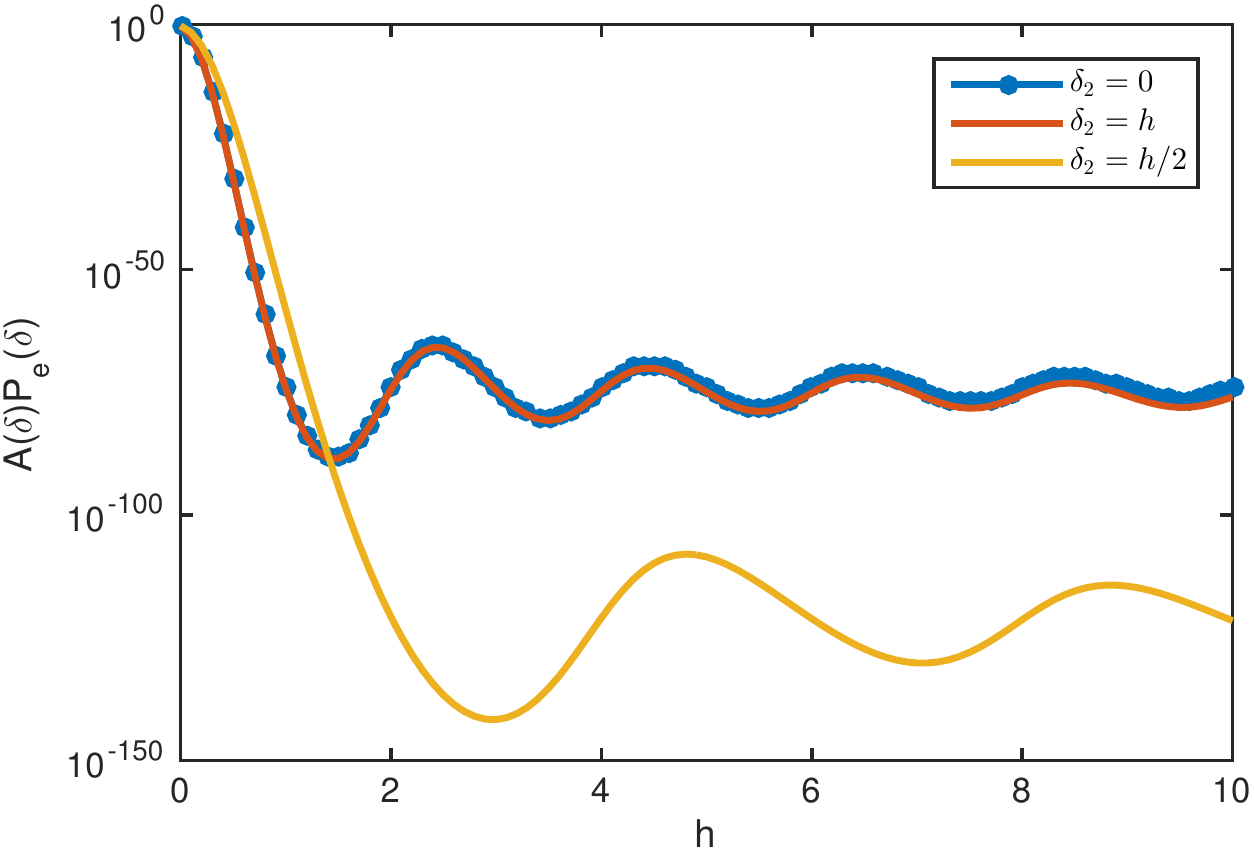}

\caption{Visualization of an example function $g(h,\delta_2)$ given by~\eqref{eq:gApp} \textbf{(left)}, and different cuts of $g(h,\delta_2)$ at $\delta_2=0, h, h/2$ \textbf{(right)}. For low $h$ values, the maximum of $g$ happens at $(h,h/2)$, while for large values it happens at $(h,0)$.}
\label{fig:gAp}
\end{figure}

An example of the function $g(\delta_1,\delta_2)$  is shown in Figure~\ref{fig:gAp}.
From Figure~\ref{fig:gAp} it is clear that there are roughly two different behaviors of $g$: one in the vicinity of $h\approx0$ where the maximum of $g(h,\delta_2)$ is obtained at $\delta_2=h/2$, while  when $h\gg0$ the maximum is obtained at $\delta_2=0$.
In what follows, we approximate the probability of error in these two different scenarios in order to reach a simplified version of the  EZZB.

\smallskip

\noindent \textbf{Small values of $h$.}
Let us note that
%\begin{align}
%a(\bdelta) + b(\bdelta)&= \frac{N_p}{2\pi}  \int_0^{W/2} \Big( \log(1+\gamma(\omega,\bdelta)) \!- \frac{\gamma(\omega,\bdelta)}{1\!+\!\gamma(\omega,\bdelta)}  \Big)\diff \omega     \nonumber \\
%&\ge  - \frac12 \frac{N_p}{2 \pi} \int_0^{W/2} \gamma(\omega,\bdelta)^2 \diff \omega,
%\end{align}
\begin{align}
a(\bdelta) + b(\bdelta) \ge  - \frac12 \frac{N_p}{2 \pi} \int_0^{W/2} \gamma(\omega,\bdelta)^2 \diff \omega,
\end{align}
which is a direct consequence of $\log(1+x) - \frac{x}{1+x} \le \frac{x^2}{2}$,  for  $x\ge0$.
Also, since $x/(1+x) \le x$, for $x\ge 0$, we have,
\begin{equation}
c(\bdelta) \mydef 2 b(\bdelta) \le \frac{N_p}{\pi} \int_0^{W/2} \gamma(\omega,\bdelta) \diff \omega.
\end{equation}

Thus, for the particular case $\bdelta =  [h,h/2]$, we have 
\begin{align}
 (K\!+\!1)^2 \!-\! T(h,\tfrac{h}{2})  &=\! 8(K\!-\!1)\sin^2(\tfrac{\omega h}{4}) \! + 4 \sin^2(\tfrac{\omega h}{2})\\
&\le \tfrac{(K\!+\!1)}{2} h^2 \omega^2.
\end{align}
The  last inequality is due the fact that $\sin^2(x) \le x^2$. This leads to an upper bound on $\gamma$, Equation~\eqref{eq:gamma},
\begin{equation}
\gamma(h,\tfrac{h}{2},\omega) \ge  \tfrac{\kappa_1 \omega^2 h^2 }{4}, \quad \text{with}\quad \kappa_1 = \tfrac{S^2  (K+1)}{2N^2 + 2(K+1)NS}. 
\end{equation}
%where
%\begin{equation}
%\kappa_1 = \tfrac{S^2  (K+1)}{2N^2 + 2(K+1)NS}. 
%\end{equation}
Thus, the probability of error can be approximately lower bounded by
{\small
\begin{align}
 P^\text{el}_\text{min} (h,\tfrac{h}{2}) &\approx e^{a(h,\tfrac{h}{2}) + b(h,\tfrac{h}{2})}  \Phi (c(h,\tfrac{h}{2})) \ge  e^{ -d^4 h^4} \Phi (c\, h)
% &\ge  e^{ -d^4 h^4} \Phi (c h),
\end{align}
}
where 
$$
d^4 =  \tfrac{W^5 N_p \kappa_1^2}{\pi \cdot 5 \cdot  2^{11}} \quad \text{and} \quad c^2 = \tfrac{ W^3 N_p \kappa_1}{\pi \cdot 3 \cdot 2^5}.
$$
Note that this lower bound on the probability of error is valid for the whole domain of $h$ and thus can be used to obtain a lower bound on the performance
of any estimator of $\btau$. For the case $\bdelta = [h,\tfrac{h}{2}]$ the function $A(\bdelta)$ given by~\eqref{eq:A} takes the form,
\begin{equation}
A(h,\tfrac{h}{2}) = (1-\tfrac{h}{D})(1-\tfrac{h}{2D})^{K-1}.
\end{equation}

Next, we have the following lower bound
%\begin{align}
%\nuC \left\{ \max_{\bdelta:\delta_1 = h} A(\bdelta) P^\text{el}_\text{min} (\bdelta) \right\} &\ge A(h,\tfrac{h}{2})  P^\text{el}_\text{min} (h,\tfrac{h}{2}) \ge f_1(h).
%\label{eq:boundSmallHAp}
%\end{align}
\begin{align}
P_A(h) &\ge A(h,\tfrac{h}{2})  P^\text{el}_\text{min} (h,\tfrac{h}{2}) \ge f_1(h),
\label{eq:boundSmallHAp}
\end{align}
where
\begin{equation}
f_1(h) = (1-\tfrac{h}{D})(1-\tfrac{h}{2D})^{K-1} e^{ -d^4 h^4} \Phi (c h).
\end{equation}
When $K$ is very large, the factor $(1-\tfrac{h}{2D})^{K-1}$  in $A(h,\tfrac{h}{2})$ significantly attenuates the probability of error. In this particular case, $\bdelta=[h,0]$ may lead to a tighter
lower bound.  Nevertheless, we will not consider this scenario.

\smallskip

\noindent \textbf{Large values of $h$.}
For large values of $h$, we will bound the performance with its value at $\bdelta=[h,0]$. According to Figure~\ref{fig:gAp}, this is the tightest path to maximize $g$ in the region $h\gg0$. 
In this case, the function
\begin{align}
g(h,0) = (1 - h/D) P^\text{el}_\text{min}(h,0),
\end{align}
oscillates outside the vicinity of 0 as illustrated in Figure~\ref{fig:gAp}. 
If 
$h_0,h_1,\ldots,h_L$ are the local maxima of $P^\text{el}_\text{min}(h,0)$, then the function
$\nuC \left\{  g(h,0) \right\}$, is tightly lower bounded by
{\small
\begin{equation}
\nuC \left\{  g(h,0) \right\} \le (1 - \tfrac{h_j}{D})  P^\text{el}_\text{min}(h_j,0) \,\,\, \text{for} \quad h_{j-1} \le h \le h_j.
\end{equation}
}
Moreover, since the function $\nuC \left\{  g(h,0) \right\}$ is non-increasing (by definition), this is true for any given set of $\{h_n\}$.
For simplicity, let us chose
%We further note that the local maxima of $P^\text{el}_\text{min}(h,0)$ are in fact the local maxima of the cross-correlation function. In a stochastic signal having flat spectrum between $[-W/2, W/2]$ these  occur approximately at 
$h_j = \frac{2\pi}{W}j$, for $j=0,1,\dotsc$.

Thus, doing similar algebraic operations as before but at $\bdelta= [h,0]$ we obtain
\begin{align}
%(K+1)^2 - T(h,0) &=  (K+1)^2 - \Big \lvert K + e^{-i \omega h} \Big \rvert^2 \nonumber\\
%&= 4K\sin^2(\omega h/2)
(K+1)^2 - T(h,0) =  4K\sin^2(\omega h/2)
\label{eq:KTh0}
\end{align}
and substituting \eqref{eq:KTh0}  in \eqref{eq:gamma}, we obtain
\begin{equation}
\gamma(h,0,\omega)\! =  \kappa_2  \sin^2 (\omega h/2 ),\,\, \text{with}\, \kappa_2 = \tfrac{S^2 K}{N^2 + (K+1)NS}.
\label{eq:gammah0}
\end{equation}
%where
%\begin{equation}
%\kappa_2 = \tfrac{S^2 K}{N^2 + (K+1)NS}.
%\end{equation}
%Note that 
%\begin{align}
%\frac{\kappa_1}{\kappa_2} = \frac{1}{2} + \frac{1}{2K}.
%\end{align}

In this case, the probability of error $P^\text{el}_\text{min}(h_j,0)$ can be tightly approximated by~\eqref{eq:peAp},
\begin{equation}
P^\text{el}_\text{min}(h_j,0) \approx e^{a(h_j,0) + b(h_j,0)}  \Phi (c(h_j,0)),
\end{equation}
where $a(h_j,0)$ and $b(h_j,0)$ are obtained by evaluating \eqref{eq:PeaAp}, \eqref{eq:PebAp} and \eqref{eq:gammah0} at $h = h_j$, respectively, obtaining
\begin{align}
a(h_j,0) %&=  -\frac{N_p}{2\pi} \int_0^{W/2} \log( 1 + \gamma(h_j,0,\omega) \diff \omega\\
 %&=-\frac{N_p}{2\pi} \int_0^{W/2} \log( 1 + \kappa_2  \sin^2 (\omega h_j/2 ) \diff \omega\\
% &=-\frac{N_p}{\pi h_j} \int_0^{W h_j/4} \log( 1 + \kappa_2  \sin^2 (\theta) \diff \theta\\
% &=-\frac{W N_p}{2 \pi^2} \int_0^{\pi/2} \log( 1 + \kappa_2  \sin^2 (\theta) \diff \theta\\
 &=-\tfrac{W N_p}{2 \pi} \log \left(\tfrac{ \sqrt{\kappa_2+1}+1}{2} \right) \mydef a,
\end{align}
and similarly
\begin{align}
b(h_j,0) %&=  \frac{N_p}{2\pi} \int_0^{W/2} \frac{\gamma(h_j,0,\omega)}{1+\gamma(h_j,0,\omega)} \diff \omega\\
% &=  \frac{N_p}{2\pi} \int_0^{W/2} \frac{\kappa_2  \sin^2 (\omega h_j/2 ) }{1+\kappa_2  \sin^2 (\omega h_j/2 ) } \diff \omega\\
% &=  \frac{N_p}{\pi h_j} \int_0^{W h_j/4} \frac{\kappa_2  \sin^2 (\theta) }{1+\kappa_2  \sin^2 (\theta) } \diff \theta\\
% &= \frac{W N_p}{2 \pi^2} \int_0^{\pi/2} \frac{\kappa_2  \sin^2 (\theta) }{1+\kappa_2  \sin^2 (\theta) } \diff \theta\\
 &=\tfrac{W N_p}{4 \pi} \tfrac{ \sqrt{\kappa_2+1} -1}{\sqrt{\kappa_2+1}}  \mydef b.
\end{align}
Note that both equations become independent of $j$. Next it follows that
%\begin{align}
$P^\text{el}_\text{min}(h_j,0) \approx e^{a + b}  \Phi (\sqrt{2b}).$
%\end{align}
Thus,
{\small
\begin{equation}
\nuC \left\{  g(h,0) \right\} \ge (1\!-\!\tfrac{h_j}{D}) e^{a+b} \Phi(\sqrt{2b})\,\, \text{for} \,\, h_j \!-\tfrac{2\pi}{W} \le h \le h_j.
\end{equation}
}
%for $h_j -\tfrac{2\pi}{W} \le g \le h_j$.
%\medskip
%
Since $1-\tfrac{h_j}{D} \ge 1-\tfrac{2\pi}{DW}$ for  $h_j \!-\! \tfrac{2 \pi}{W} \le h \le h_j$, it follows that
%\begin{align}
%\nuC \left\{ \max_{\bdelta:\delta_1 = h} A(\bdelta) P^\text{el}_\text{min} (\bdelta) \right\} &\ge \nuC \left\{  g(h,0) \right\} \ge f_2(h), 
%\label{eq:boundLargeHAp}
%\end{align}
\begin{align}
P_A(h) &\ge \nuC \left\{  g(h,0) \right\} \ge f_2(h), 
\label{eq:boundLargeHAp}
\end{align}
where 
\begin{equation}
%f_2(h) =  (D- \tfrac{2 \pi}{W} - h)^+ e^{a+b} \phi(\sqrt{2b}).
f_2(h) =  \max(1- \tfrac{2 \pi}{DW} - \tfrac{h}{D},0) e^{a+b} \phi(\sqrt{2b}).
%
% f^+(x)=max(f(x),0)
 %
 %
\end{equation}

\smallskip

\noindent \textbf{Final Bound.}
To get the final bound we merge the two previous lower-bounds, \eqref{eq:boundSmallHAp} and \eqref{eq:boundLargeHAp}, 
into a single lower bound,
%\begin{align}
%\nuC \left\{ \max_{\bdelta:\delta_1 = h} A(\bdelta) P^\text{el}_\text{min} (\bdelta) \right\}  \ge \max \left(f_1(h) , f_2(h) \right).
%\end{align}
\begin{align}
P_A(h)  \ge \max \left(f_1(h) , f_2(h) \right).
\end{align}

To further simplify the lower bound we can split the domain of integration of $h$, that is $[0,D]$, and make each one valid in a region.
%In general this will not be the optimal solution, since we cannot determine analytically the best partition for the optimal assignment. 
%
Let $h^{\ast} = \sqrt{2b}/c$.  This point is close to the intersection of $f_1(h)$ and $f_2(h)$. 
Thus,
%\begin{align}
%\nuC \left\{ \max_{\bdelta:\delta_1 = h} A(\bdelta) P^\text{el}_\text{min} (\bdelta) \right\}  \ge 
%\begin{cases}
%f_1(h)  & \text{if } 0 \le h < h^\ast ,\\
%f_2(h)  & \text{if }  h^\ast \le h.\\
%\end{cases}
%\label{eq:boundAllHAp}
%\end{align}
\begin{align}
P_A(h)  \ge 
\begin{cases}
f_1(h)  & \text{if } 0 \le h < h^\ast ,\\
f_2(h)  & \text{if }  h^\ast \le h.\\
\end{cases}
\label{eq:boundAllHAp}
\end{align}

Substituting \eqref{eq:boundAllHAp} in \eqref{eq:ezzb1Ap} one obtains a lower bound on the mean square error,
\begin{align}
\epsilon^2_1 \ge \int_0^{h^\ast} h f_1(h)  \diff h  +   \int_{h^\ast}^D h f_2(h)  \diff h.
\label{eq:error1Ap}
\end{align}

The first term in \eqref{eq:error1Ap} can be lower bounded by
%{\small
%\begin{align}
%&\int_0^{h^\ast} h f_1(h)  \diff h \nonumber \\
%%&=  \int_0^{h^\ast} h(1-h/D)(1-h/{2D})^{K-1} e^{ -d^4 h^4} \Phi (c h) \diff h \nonumber \\
%& \ge (1\!-\!\tfrac{h^\ast}{D})(1\!-\!\tfrac{h^\ast}{2D})^{K\!-\!1}  \int_0^{h^\ast} \! he^{ -d^4 h^4} \Phi (c\, h) \diff h. \label{eq:Peterm1Ap}
%\end{align}
%}
{\small
\begin{equation}
\int_0^{h^\ast} h f_1(h)  \diff h \ge (1\!-\!\tfrac{h^\ast}{D})(1\!-\!\tfrac{h^\ast}{2D})^{K\!-\!1}  \int_0^{h^\ast} \! he^{ -d^4 h^4} \Phi (c\, h) \diff h. \label{eq:Peterm1Ap}
\end{equation}
}

When $K$ is not very large, $(1-\tfrac{h^\ast}{D})(1-\tfrac{h^\ast}{2D})^{K-1}  \approx 1$. Then,
%\begin{align}
% \int_0^{h^\ast} he^{ -d^4 h^4} \Phi (c h) \diff h &= \frac{1}{c^2} \int_0^{c h^\ast} x e^{- (d/c)^4 h^4} \Phi(h) \diff h \nonumber\\
%  &= \frac{1}{c^2} \int_0^{\sqrt{2b}} h e^{-\frac{9 \pi h^4 }{10 WN_p}} \Phi(h) \diff h. \label{eq:term1}
%\end{align}
{\small
\begin{align}
 \int_0^{h^\ast} he^{ -d^4 h^4} \Phi (c\, h) \diff h &= \frac{1}{c^2} \int_0^{\sqrt{2b}} h e^{-\frac{9 \pi h^4 }{10 WN_p}} \Phi(h) \diff h. \label{eq:term1}
\end{align}
}
The second term in \eqref{eq:error1Ap} can be (approx.) lower bounded by
{\small
 \begin{align}
  &\int_{h^\ast}^D h f_2(h)  \diff h = \int_{h^\ast}^D \max(1\!-\! \tfrac{2 \pi}{WD} \!-\! \tfrac{h}{D},0) e^{a+b} \phi(\sqrt{2b})  h \diff h \nonumber\\
 % &\quad\quad \int_{h^\ast}^D h(1\!-\! \tfrac{2 \pi}{WD} \!-\! \tfrac{h}{D})^+ e^{a+b} \phi(\sqrt{2b})  \diff h \nonumber\\
  &\ge   e^{a+b} \phi(\sqrt{2b})\int_{4 \sqrt{3}/W}^{D-2\pi/W} \max(1- \tfrac{2 \pi}{WD} - \tfrac{h}{D},0) h \diff h \nonumber\\
% &= \frac{D^2}{6} + \frac{64 \sqrt{3}}{D W^3} + \frac{48 \pi}{D W^3} - \frac{4 \pi^3}{3 D W^3} - \frac{24}{W^2} +  \frac{2 \pi^2}{W^2} - \frac{D \pi}{W}\\
 % &=  e^{a+b} \phi(\sqrt{2b})  \Big( \frac{D^2}{6} \!+\! \tfrac{64 \sqrt{3} \!+\! 48\pi \!-\! 4\pi^3}{D W^3}  \!-\! \frac{24 \!-\! 2\pi^2}{W^2} \!-\! \frac{D \pi}{W} \Big) \nonumber\\
  % &\ge e^{a+b} \phi(\sqrt{2b})  \Big(\frac{D^2}{6}   \!-\! \tfrac{24 - 2\pi^2}{W^2} \!-\! \frac{D \pi}{W} \Big) \nonumber \\
   &\approx  \tfrac{D^2}{6}e^{a+b} \phi(\sqrt{2b})     \label{eq:Peterm2Ap}.
\end{align}
}
From  \eqref{eq:error1Ap},  \eqref{eq:term1} and \eqref{eq:Peterm2Ap}, we get the EZZB bound in~\eqref{eq:ezzbWS}.

\bibliographystyle{IEEEtran}
\bibliography{references}
%

\end{document}